\documentclass{article}

\usepackage{PRIMEarxiv}

\usepackage[utf8]{inputenc} 
\usepackage[T1]{fontenc}    
\usepackage{url}            
\usepackage{booktabs}       
\usepackage{amsfonts}       
\usepackage{nicefrac}       
\usepackage{microtype}      
\usepackage{lipsum}
\usepackage{fancyhdr}       
\usepackage{graphicx}       
\graphicspath{{media/}}     
\usepackage{amsmath} 

\usepackage{array}
\usepackage{todonotes}
\usepackage[numbers]{natbib}
\usepackage{authblk}  
\usepackage{hyperref}
\usepackage{xcolor}  

\usepackage{lineno}

\title{

AI-Driven Day-to-Day Route Choice

}

\newcommand{\correspondingauthor}{\thanks{Corresponding author}}
\author[1]{Leizhen Wang}
\author[1$^*$]{Peibo Duan}  
\author[2]{Zhengbing He}
\author[3]{Cheng Lyu}
\author[4]{Xin Chen}
\author[5]{Nan Zheng}
\author[6]{Li Yao}
\author[7\correspondingauthor]{Zhenliang Ma}  

\affil[1]{Department of Data Science and Artificial Intelligence, Monash University, Melbourne, Australia}
\affil[2]{Laboratory for Information \& Decision Systems, Massachusetts Institute of Technology, Cambridge, United States}
\affil[3]{Chair of Transportation Systems Engineering, Technical University of Munich, Munich, Germany}
\affil[4]{School of Civil Engineering, The University of Queensland, Brisbane, Australia}
\affil[5]{Department of Civil Engineering, Monash University, Melbourne, Australia}
\affil[6]{School of Computer Science and Engineering, Southeast University, Nanjing, People's Republic of China}
\affil[7]{Department of Civil and Architectural Engineering, KTH Royal Institute of Technology, Stockholm, Sweden}

\begin{document}
{\setlength{\baselineskip}{1.2\baselineskip}

\maketitle

\begin{abstract}

Understanding travelers' route choices can help policymakers devise optimal operational and planning strategies for both normal and abnormal circumstances. However, existing choice modeling methods often rely on predefined assumptions and struggle to capture the dynamic and adaptive nature of travel behavior. Recently, Large Language Models (LLMs) have emerged as a promising alternative, demonstrating remarkable ability to replicate human-like behaviors across various fields. Despite this potential, their capacity to accurately simulate human route choice behavior in transportation contexts remains doubtful. To satisfy this curiosity, this paper investigates the potential of LLMs for route choice modeling by introducing an LLM-empowered agent, "LLMTraveler." This agent integrates an LLM as its core, equipped with a memory system that learns from past experiences and makes decisions by balancing retrieved data and personality traits. The study systematically evaluates the LLMTraveler's ability to replicate human-like decision-making through two stages of day-to-day (DTD) congestion games: (1) analyzing its route-switching behavior in single origin-destination (OD) pair scenarios, where it demonstrates patterns that align with laboratory data but cannot be fully explained by traditional models, and (2) testing its capacity to model adaptive learning behaviors in multi-OD scenarios on the Ortuzar and Willumsen (OW) network, producing results comparable to Multinomial Logit (MNL) and Reinforcement Learning (RL) models. Additionally, the study assesses lightweight, open-source LLMs, highlighting their effectiveness in route choice simulation and their potential as cost-effective alternatives to more advanced closed-source models. These experiments demonstrate that the framework can partially replicate human-like decision-making in route choice while providing natural language explanations for its decisions. This capability offers valuable insights for transportation policymaking, such as simulating traveler responses to new policies or changes in the network. The code for this paper is open-source and available at: \href{https://github.com/georgewanglz2019/LLMTraveler}{https://github.com/georgewanglz2019/LLMTraveler}.

\end{abstract}

\keywords{Route Choice \and Large Language Models \and Congestion Game \and Generative Agent}



\newpage


\section{Introduction}

Understanding individual travel behaviors is critical for developing efficient and sustainable transportation systems. Travel behavioral analysis aims to capture the decision-making process of individual travel execution, including travel route choice, travel mode choice, departure time choice, and trip purpose. Among these choices, modeling route choice not only helps analyze and understand travelers' behaviors, but also constitutes the essential part of traffic assignment methods \citep{prato2009route}. Specifically, it enables the evaluation of travelers' perceptions of route characteristics, the forecasting of behavior in hypothetical scenarios, the prediction of future traffic dynamics on transportation networks, and the understanding of travelers' responses to travel information.

Real-world route choice is complex because of the inherent difficulties in accurately representing human behavior, travelers' limited knowledge of network composition, uncertainties in perceptions of route characteristics, and the lack of precise information about travelers' preferences \citep{prato2009route}. To overcome these limitations, DTD traffic dynamics have attracted significant attention since they focus on drivers' dynamic shifts in route choices and the evolution of traffic flow over time, rather than merely static equilibrium states. DTD models are flexible to incorporate diverse behavioral rules such as forecasting \citep{he2012modeling, xiao2016physics}, bounded rationality \citep{guo2011bounded, ye2017rational}, decision-making based on prospects \citep{xu2011decision, wang2013combined}, marginal utility effects \citep{kumar2015day, he2016marginal}, and social interactions \citep{wei2016day}. Despite these advantages identified in \citep{smith1984stability} and \citep{horowitz1984stability}, DTD models still struggle to accurately reflect the observed fluctuations in traffic dynamics, particularly the persistent deviations around User Equilibrium (UE) noted in empirical studies \citep{iida1992experimental, selten2007commuters, meneguzzer2013day}.

To better understand traffic dynamics, Agent-Based Modeling (ABM) offers a promising alternative. It excels at simulating individual behaviors, revealing complex patterns from local interactions and decisions of individual agents \citep{zou2013dynamic}. The ABM was first introduced into the route choice model for the study of the impact of information from different sources on travelers and to model their responses and learning behaviors \citep{jha1998perception, nakayama2000route, dia2002agent, rossetti2005dynamic}. In recent years, many studies have also explored the use of model-free RL-based methods to model the learning behaviors of selfish agents in congestion games \citep{ramos2018analysing, mao2018reinforcement, zhou2020reinforcement, shou2022multi}. While these advances have contributed to the development of route choice modeling, both ABM and RL-based methods face limitations in modeling the dynamic and adaptive nature of traveler behavior. ABM relies on predefined decision rules that are typically static, making it challenging to reflect evolving decision-making processes. RL, while capable of learning strategies from interactions, depends heavily on training within fixed distributions, limiting its ability to generalize to new or unforeseen conditions. As a result, these models often struggle to adapt to dynamic and unexpected scenarios without retraining or manual adjustments.

Recently, the appearance of LLMs, such as Generative Pre-Trained Transformers 4 (GPT-4) \citep{achiam2023gpt}, marks a significant milestone in machine learning, showing great potential in natural language processing and text generation. Trained on extensive datasets, LLMs can reflect a broad spectrum of human behaviors, providing profound insights into complex decision-making processes. Notably, \citep{mei2024turing} shows that GPT-4 has passed a Turing test, demonstrating behavior and personality traits that are statistically indistinguishable from a diverse human sample spanning over 50 countries. The application of LLMs in simulating human behavior has seen rapid progress, including various fields such as social science, gaming, psychology, economics, and policy-making \citep{guo2024large, park2023generative, gao2024large}. These studies reveal highly human-like behaviors and display complex phenomena similar to real-world scenarios within various systems.

A key advantage of LLMs is their adaptability. Unlike traditional models that rely on static assumptions and predefined rules, LLMs can well comprehend human common sense and they are few-shot learners that can efficiently adapt to new information with instructions or limited examples \citep{brown2020language, wang2024survey}. This high level of adaptability allows LLMs to respond dynamically to changes in the environment and adjust their behavior accordingly \citep{gao2024large}.  Integrating LLMs into route choice modeling can effectively address the critical challenges faced by traditional methods. Moreover, LLMs offer enhanced explainability by generating natural language explanations for their decisions, providing transparency into their decision-making process. 

Although LLMs have been widely applied in various research fields, their ability to accurately simulate human route choice behavior in transportation environments remains doubtful. LLMs are primarily designed to predict the next token in a sequence, which raises the challenge of how to effectively design prompts that allow the model to understand and simulate the decision-making process in real-world route choice scenarios. Moreover, human travelers in urban networks learn and adapt their behavior over time, based on past experiences. This dynamic, evolving nature of decision-making presents an additional challenge in integrating LLMs to replicate such adaptive behaviors. In addition, while powerful closed-source LLMs such as GPT-4 offer substantial capabilities, their high costs can be prohibitive. This raises the question of whether lightweight, open-source LLMs can be effectively utilized for route choice modeling. Exploring this possibility could provide a cost-effective and accessible alternative, broadening the applicability of LLMs in transportation research and practice. This paper aims to address these challenges.

To best of our knowledge, this study is the first to apply LLMs to model and interpret travelers' route choice behaviors in urban mobility systems. The paper makes the following contributions:

\begin{itemize}

\item Introducing "LLMTraveler," an LLM-empowered agent for route choice modeling, which integrates advanced prompt engineering and a memory system to simulate human travel decisions, setting a new standard for using LLMs in travel behavior simulation.
\item Evaluating the performance of the proposed LLMTraveler across single OD and multi-OD pair scenarios, comparing its behavior with laboratory data, traditional theoretical models, and reinforcement learning-based approaches.
\item Assessing the feasibility of using lightweight, open-source LLMs for route choice simulation, demonstrating their potential as cost-effective alternatives to more advanced closed-source models.

\end{itemize}

These contributions offer new insights into travel behavior and provide transportation policymakers with a novel tool to design effective traffic management strategies and predict user responses. For instance, in the absence of historical data, governments could use LLM-powered agents to simulate the impact of new policies or changes to the transportation network. 

The remainder of this paper is organized as follows. Section~\ref{sec:methods} presents the methodology behind the proposed LLMTraveler agent. Section~\ref{sec:eval-singleOD} and \ref{sec:eval-multiOD} evaluate the agent’s performance in single and multi-OD network scenarios, respectively, comparing its behavior with laboratory data and traditional models. Finally, Section~\ref{sec:conclusion} concludes the paper, summarizing the key findings and suggesting directions for future research.

\section{Method}
\label{sec:methods}
\subsection{Preliminaries}
LLMs are typically built on the Transformer \citep{vaswani2017attention} architecture, which employs self-attention mechanisms to effectively model long-range dependencies in text. Modern LLMs contain hundreds of millions or even billions of parameters, significantly enhancing their performance across a wide range of tasks, from natural language understanding to text generation.

A notable advancement in LLM capabilities is the concept of in-context learning (ICL), formally introduced by GPT-3 \citep{brown2020language}. With ICL, a language model can generate expected outputs based on natural language instructions and task demonstrations provided within the prompt, without requiring additional training or parameter updates. This ability enables LLMs particularly useful for simulating human route choice decisions, as a well-designed prompt can guide the model to generate accurate and relevant answers based on the given information: 
\begin{equation}
    \text{seq}^* = \arg\max_{\text{seq}} P(seq \mid \text{Context}; \theta^*)
\end{equation}
where \(\text{seq}\) is the output sequence. \(\text{seq}^*\) is the output sequence with the highest probability. \(\theta^*\) are the optimized model parameters obtained through pre-training. \(\text{Context}\) represents all natural language instructions and/or several task demonstrations used to guide the LLM in generating the desired output sequence. 
Based on the ICL, LLMs can be used as an agent's brain to simulate traveler's route choice behavior by designing effective prompts.

\subsection{LLM-empowered traveler agent}
In this subsection, the proposed method, LLMTraveler, an LLM-empowered intelligent agent for travel behavior modeling, is presented. Figure~\ref{fig:TA_framework} shows its framework. First, the decision-making process of the LLMTraveler is discussed, followed by a detailed explanation of its prompt design.

\begin{figure}
  \centering
  \includegraphics[width=0.55\textwidth]{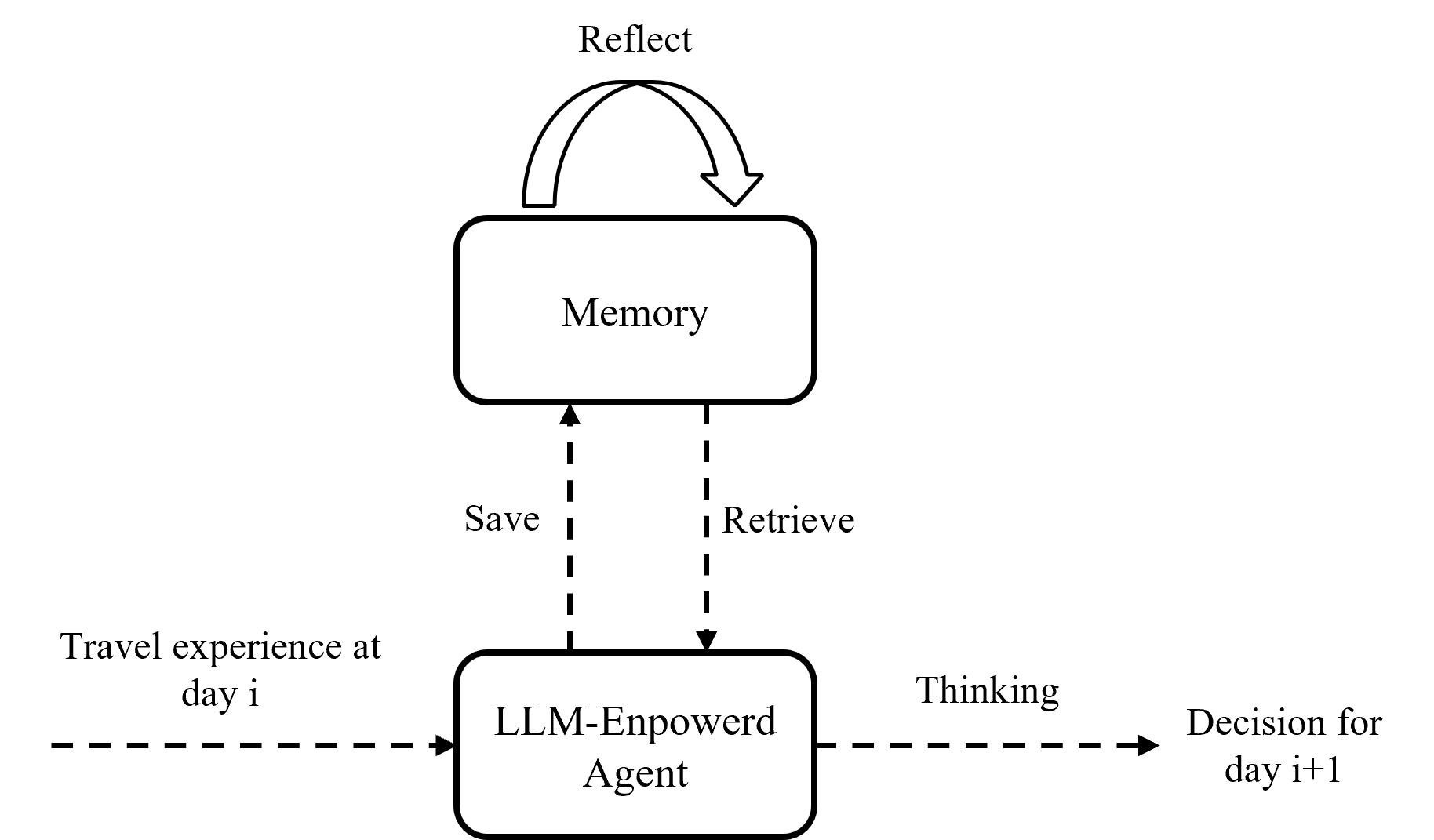}
  \caption{Framework of LLMTraveler}
  \label{fig:TA_framework}
\end{figure}

\subsubsection{Decision Process}

In the proposed framework, the decision process of each LLMTraveler is designed to mimic human-like decision-making in route choices. This involves the following three steps:

\textbf{Step 1: Reception and update of travel cost information}.
At the end of day \( i \), each LLMTraveler receives travel cost information for the chosen route or all available routes. The agent updates its memory with this information, storing travel cost data along with other relevant details such as traffic conditions and route-specific costs. This comprehensive data collection helps in building a robust memory of travel experiences.

\textbf{Step 2: Reflection on route choices}.
The agent reflects on its past route choices based on the accumulated experience. For the experiments, this reflection involves calculating the Exponential Weighted Moving Average Travel Time (EWMATT) for each route to prioritize recent experiences while still considering historically experienced data. The formula used is:
\begin{equation}
    \text{EWMATT}_t = \omega \cdot T_t + (1 - \omega) \cdot \text{EWMATT}_{t-1}
\end{equation}
where \(\text{EWMATT}_t\) is the EWMATT after the \( t \)-th update, \(\omega\) is the smoothing factor, \(T_t\) is the \( t \)-th observed travel time, and \(\text{EWMATT}_{t-1}\) is the EWMATT after the \( (t-1) \)-th update.

This reflection can also be implemented through other methods, such as having the LLM perform self-refinement \citep{madaan2024self}.

\textbf{Step 3: Decision making using LLMs}.
The LLMTraveler retrieves data from memory (e.g., EWMATT and the chosen times of each route) and uses this information to construct a prompt. This prompt is then fed into the LLM, which processes it and generates a route choice decision for day \( i+1 \).

\subsubsection{Prompt Design}

The context-inclusive prompts, which integrate the traveler profile, task description, travel experiences, thinking guidance, output format, and reasoning, are shown in Figure~\ref{fig:prompt}. This comprehensive approach aims to enhance the LLMTraveler's capabilities by building upon existing prompting strategies. This section details the prompt design for the LLMTraveler, focusing on the key components necessary for accurate and realistic route choice modeling.

\begin{figure}
  \centering
  \includegraphics[width=0.7\textwidth]{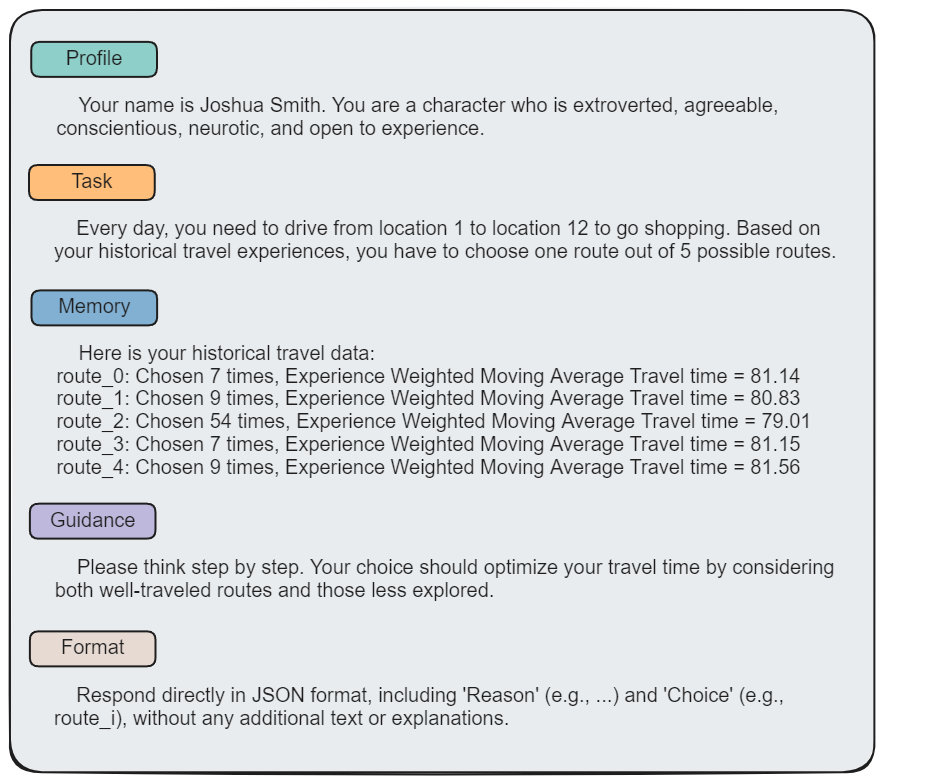}
  \caption{Example prompt template}
  \label{fig:prompt}
\end{figure}

\textbf{Traveler Profile}.
Individual attributes significantly influence travel behavior \citep{mwale2022factors, parr2016differential}. The prompt design includes a detailed profile of the traveler, such as a randomly generated name, personality traits (e.g., extroverted, agreeable, conscientious, neurotic, and open to experience), selfish or not, demographic and socioeconomic characteristics, and risk preferences. This profile helps the LLMs understand the decision-making context from the traveler's perspective, ensuring that the agent's decisions are personalized and contextually relevant.

\textbf{Task description}.
The task is to simulate route choice behavior. Each day, the agent needs to decide on a route based on historical travel experiences. The task description specifies the daily goal and outlines the available route options and relevant travel data. This setup provides the LLMs with a clear understanding of the agent's objectives and constraints.

\textbf{Travel experiences}.
Travel behavior is largely influenced by historical experiences \citep{vacca2019should}. The prompt includes the retrieved data, such as the chosen times and EWMATT of each route. This module allows the LLMs to incorporate past experiences into the decision-making process, ensuring that decisions are informed by a comprehensive historical context.

\textbf{Thinking guidance}.
To guide the LLMs' reasoning, the prompt includes a guidance section instructing the agent to "think step-by-step" (zero-shot chain-of-thought strategy \citep{wei2022chain, zhang2022automatic}, optimizing route choice by "considering both well-traveled routes and less explored options." This guidance leverages domain knowledge and common sense to enhance the LLMs' reasoning capabilities, helping them to balance various factors effectively.

\textbf{Output format}.
The prompt format is designed to ensure clarity and ease of interpretation for the LLM. It specifies that the agent's response should be in JSON format, including both the selected route and the reasoning behind the choice. This structured format enables efficient processing and simplifies the analysis of the LLMs' output, ensuring the results are easily extractable and interpretable.

\textbf{Interpretation and output}.
The output from the LLMs includes two parts: the route choice and the reasoning behind the choice. By first asking for the reason and then the decision, the prompt not only improves the interpretability of the model but also enhances its reasoning ability. This approach encourages the LLMs to thoroughly process the information before making decisions, which improves the model's reasoning performance and provides deeper insights into the decision-making process \citep{yao2022react}.

\subsection{DTD Route Choice Modeling Framework}

Figure~\ref{fig:D2D} shows the DTD route choice modeling framework, which is designed to simulate the iterative decision-making process of travelers within a transportation network over multiple days. This iterative approach provides valuable insights into the evolution of route choices over time.

\begin{figure}
  \centering
  \includegraphics[width=0.98\textwidth]{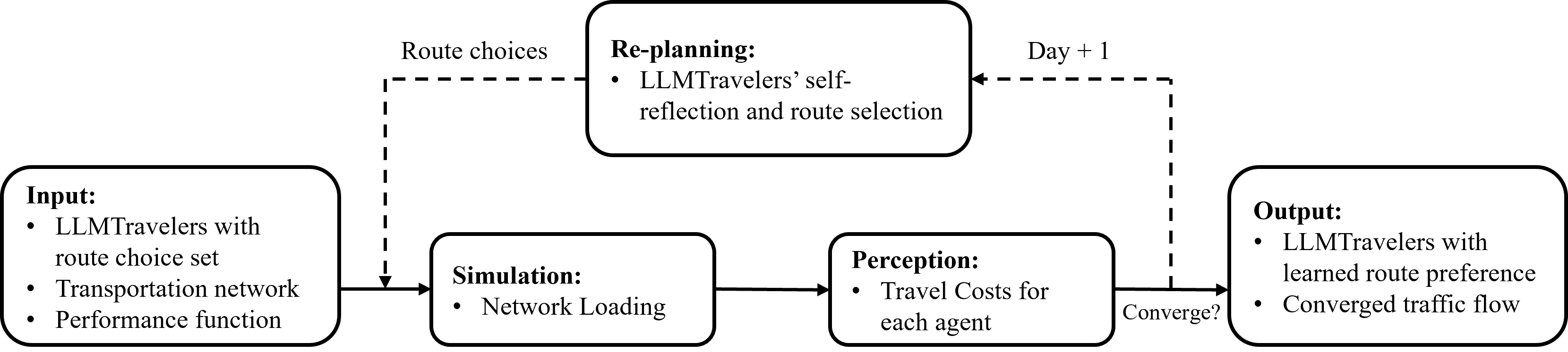}
  \caption{DTD route choice modeling framework}
  \label{fig:D2D}
\end{figure}

\textbf{Input}.
The input for this framework consists of three main components:
\begin{itemize}
    \item \textbf{Transportation Network}. 
    This includes nodes, edges, and their connectivity. The network structure defines the possible paths that travelers can take.
    
    \item \textbf{Demand (LLMTravelers)}. 
    This involves the OD pairs and the number of travelers between these pairs. In the framework, each traveler or group of travelers is considered a LLMTraveler, complete with a specific profile. For each OD pair, \(k\) alternative routes are computed using the \(k\)-shortest path algorithm \citep{yen1970algorithm}, providing multiple options for each LLMTraveler.
    
    \item \textbf{Link Performance Function}. 
    This function calculates the travel time for each link based on the traffic volume. It can be linear or non-linear, depending on the specific characteristics of the transportation network. In the experiments, a linear performance function is used. Despite its simplicity, the linear cost function effectively captures the essential relationship between travel time and traffic flow \citep{qi2023investigating}:
    \begin{equation}
        \text{tt}_t^k = t_0^k + \sigma \cdot x_t^k
    \end{equation}

    where \(\text{tt}_t^k\) is the travel cost (e.g., travel time) for link \(k\) at day \(t\), \(t_0^k\) is the free flow travel cost for link \(k\), \(x_t^k\) is the traffic volume of link \(k\) at day \(t\), and \(\sigma\) is a constant that scales the impact of the traffic volume.
\end{itemize}

\textbf{Simulation Loop}.
The simulation loop operates over a predefined number of days, representing the iterative nature of travelers' route choice decisions.
\begin{itemize}
    \item \textbf{Route Choice}. 
    For day \(i\), each LLMTraveler selects a route from their available alternatives based on their own experiences and information.
    
    \item \textbf{Simulation and Perception}. 
    After all LLMTravelers have chosen their routes, the network is loaded with these selected routes. This step involves calculating the travel times for each link using the performance function for the day. The travel cost for all chosen routes, along with other relevant information such as traffic conditions and route-specific costs, is perceived by each LLMTraveler.
    
    \item \textbf{Re-planning}. 
    Following the network loading, each LLMTraveler updates their experiences based on the travel cost of the day. This includes storing the travel times and relevant information, which will influence their future route choices. The memory of each agent is updated to reflect the day's travel data, ensuring continuous learning and adaptation. They will use this updated information to make route decisions for day \(i+1\).
\end{itemize}

\textbf{Output}.
After sufficient days/iterations or upon meeting certain convergence criteria, the final day's traffic flow is returned. LLMTravelers, having accumulated extensive travel experiences, will exhibit established route preferences, providing insights into the long-term equilibrium state of the transportation network.

\section{Evaluation of route choice behaviors in a single OD pair network}
\label{sec:eval-singleOD}

This section evaluates the route choice behavior regularities of the proposed LLMTraveler in a single OD pair congestion game. The agent's behavior patterns are compared with laboratory data and traditional theoretical models.

\subsection{Experiment Settings}
Figure~\ref{fig:OneOD} shows the single OD pair network, which is commonly used in behavior experimental studies \citep{iida1992experimental, selten2007commuters, qi2023investigating, zhang2018cumulative, meneguzzer2019contrarians, zhao2016experiment}. In these scenarios, a fixed number of travelers (e.g., 16 in this experiment) commute from the same origin to the same destination with two route choices every morning. Travel time on each route was assumed to increase with route flow. This experimental setup closely follows the methodology outlined in \citep{qi2023investigating}, ensuring consistency with their approach.

\begin{figure}
  \centering
  \includegraphics[width=0.2\textwidth]{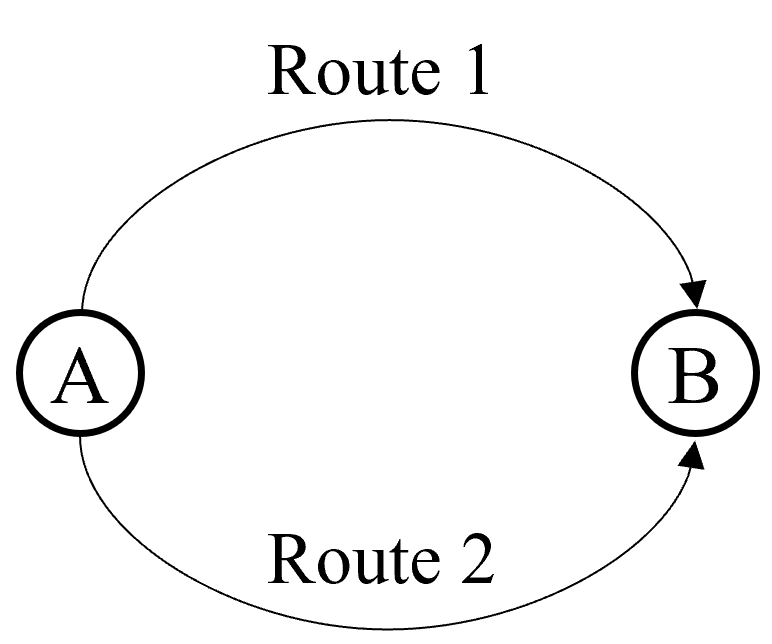}
  \caption{Single OD pair network}
  \label{fig:OneOD}
\end{figure}

The five scenarios illustrated in Figure~\ref{fig:OneOD}, with detailed cost functions provided in Table~\ref{tab:one_od_scenario_setting}, are designed as follows:   
\begin{itemize}
    \item \textbf{Scenario 1}. 
    This serves as the baseline, featuring a symmetric two-route network.
    
    \item \textbf{Scenarios 2–5}. 
    These extend Scenario 1 to asymmetric two-route networks with varying cost functions, designed to investigate travelers' route choice behaviors under different cost feedback conditions. The cost functions are based on the previous research in~\citep{selten2007commuters, qi2023investigating}.
\end{itemize}

Additionally, the LLM temperature is set to 0 by default, with
typical values ranging between 0 and 2. The smoothing factor \(\omega\) used to calculate EWMATT was set to 0.2 in the experiments. Each scenario was simulated for 100 days per run, with a total of three runs conducted to ensure robust results.

\begin{table}[h]
    \centering
    \small 
    \caption{Scenario settings}
    \label{tab:one_od_scenario_setting}
    \begin{tabular}{c >{\centering\arraybackslash}m{4cm} c}
        \toprule
        \textbf{Scenario} & \textbf{Cost function\textsuperscript{a}} & \textbf{DUE\textsuperscript{b}} \\
        \midrule
        1 & 
        $\begin{aligned} 
            c_1 &= 6 + 2f_1 \\ 
            c_2 &= 6 + 2f_2 
        \end{aligned}$ & 8, 8 \\
        \midrule
        2 & 
        $\begin{aligned} 
            c_1 &= 10 + 4f_1 \\ 
            c_2 &= 24 + 6f_2 
        \end{aligned}$ & 11, 5 \\
        \midrule
        3 & 
        $\begin{aligned} 
            c_1 &= 5 + 2f_1 \\ 
            c_2 &= 12 + 3f_2 
        \end{aligned}$ & 11, 5 \\
        \midrule
        4 & 
        $\begin{aligned} 
            c_1 &= 12 + 4f_1 \\ 
            c_2 &= 24 + 6f_2 
        \end{aligned}$ & 10.8, 5.2 \\
        \midrule
        5 & 
        $\begin{aligned} 
            c_1 &= 6 + 2f_1 \\ 
            c_2 &= 12 + 3f_2 
        \end{aligned}$ & 10.8, 5.2 \\
        \bottomrule
    \end{tabular}
    \vspace{0.5em}
    
    \textsuperscript{a} $c_i$ and $f_i$ are the cost and flow of route $i$, respectively. \\
    \textsuperscript{b} Dynamic User Equilibrium (DUE) flow assignment; the unit is "traveler".
\end{table}

\subsection{Evaluation metrics}
The average switching rate is used for evaluation. To provide clarity, the switching rate is defined as follows:

\textbf{Switching rate \citep{qi2023investigating, qi2024investigating}}. 
This metric measures how travelers switch routes over time. Consider an OD pair with several feasible routes. The switching rate \( p_{ij}^t \) is defined as the proportion of travelers switching from route \( i \) to route \( j \) between time \( t \) and \( t+1 \). For scenarios with only two routes, the proportion of travelers remaining on their previously chosen route \( i \) is given by \( p_{ii}^t = 1 - p_{ij}^t \).

\textbf{Average switching rate \citep{qi2023investigating, qi2024investigating}}. 
To analyze the observed switching rates and identify gaps between the LLMTraveler, existing theories and laboratory observations, the relationship between average switching rates and cost combinations is investigated. The \textit{average switching rate} $\bar{p}_{ij}(\vec{c})$ is employed as an intuitive indicator, calculated as follows:
\begin{equation}
    \bar{p}_{ij}(\vec{c}) = \frac{\sum_{t \in T(\vec{c})} p_{ij}^t}{|T(\vec{c})|}
\end{equation}
where $\vec{c}$ represents a specific cost combination, and $T(\vec{c})$ denotes the set of days during which the costs of all routes are equal to $\vec{c}$. The average switching rate serves as a reflection of the population's average behavior, effectively minimizing the influence of individual heterogeneity and stochastic factors.

\subsection{Compared Methods}
\label{subsec:eval_compared_method}

The performance of the proposed method is compared with the following three baseline methods:

\textbf{Perfectly Rational Choice (PRC) model}. 
This model follows an intuitive, rational, and individual-level route choice rule. Travelers switch their routes only if there is a less costly alternative unless all travelers were already on the least costly routes the previous day \citep{zhang2001equivalence}. In the experiments, every agent is assumed to select routes based on the principle of cost minimization, using EWMATT as the cost measure.

\textbf{MNL model}. 
This model assumes that travelers select routes based on perceived utility maximization rather than actual utility~\citep{daganzo1977stochastic}. In the experiments, the utility is defined as a hyperparameter \( \alpha \) multiplied by the negative EWMATT. For instance, "MNL-0.3" represents an MNL model with \( \alpha \) set to 0.3, while "MNL" denotes the default model with \( \alpha \) set to 1.

\textbf{LLMTraveler}. 
This is the proposed method. Table~\ref{tab:llm_summary} provides a summary of the LLM models used, including their corresponding codes, LLMs' model names, parameter counts, and whether they are open or closed-source. In the experiments, the LLMTraveler is named according to the LLM serving as its core. For example, "LLM-gpt35" refers to a LLMTraveler using "GPT-3.5-Turbo-1106" as its LLM core. The selected LLM models cover a range of sizes, from smaller models such as "llama-3.1-8b" to larger models like "GPT-4o". The selection includes both open-source and closed-source models, with parameter counts ranging from 7 billion to over 175 billion.

\begin{table}[h]
    \centering
    \small 
    \caption{Summary of LLMTraveler used in experiments}
    \label{tab:llm_summary}
    \begin{tabular}{cccc}
        \toprule
        \textbf{Code} & \textbf{LLM} & \textbf{Parameter Count} & \textbf{Source} \\
        \midrule
        LLM-gpt4o & GPT-4o-2024-05-13 \cite{achiam2023gpt} & >175 billion & Closed-source \\
        LLM-gpt35 & GPT-3.5-Turbo-1106 \cite{brown2020language} & 175 billion & Closed-source \\
        LLM-llama-3.1-70b & llama-3.1-70b-instruct \cite{dubey2024llama} & 70 billion & Open-source \\
        LLM-yi-medium & yi-medium \cite{young2024yi} & 34 billion & Open-source \\
        LLM-llama-3.1-8b & llama-3.1-8b-instruct \cite{dubey2024llama} & 8 billion & Open-source \\
        \bottomrule
    \end{tabular}
\end{table}

\subsection{Experiment Results}

\subsubsection{Route choice examples and rationales of different LLMTravelers}

Table~\ref{tab:binary_choice_and_reason_examples} presents some examples of the route choice behavior of various LLM-based agents, with decisions influenced by their profiles, historical data, and randomness. For instance, LLM-gpt4o (row 1) demonstrates a preference for randomness in decision-making, selecting "route 1" despite "route 2" has higher historical performance, as shown in its reasoning ("route 2 has a slightly lower..."). On the other hand, LLM-gpt35 (row 2) adopts a more cautious approach, choosing "route 2" due to its lower average travel time, reflecting its risk-averse personality. Similarly, LLM-llama3.1-70b (row 4) balances historical performance with a probabilistic approach, opting for "route 2" but introducing randomness. In contrast, LLM-llama3.1-8b (row 5) factors in minimal differences in travel time and the impact of others' choices. Furthermore, LLM-yi-medium (row 3), with its risk-neutral and extroverted traits, chooses "route 1" despite "route 2" offering a shorter travel time on average, highlighting its inclination toward balancing the historical data with an element of unpredictability. These varied strategies highlight the agents' ability to mimic human decision-making, blending rational analysis with uncertainty, making them suitable for simulating route choice in transportation studies.

\begin{table}[h]
    \centering
    \caption{Route choice and reason examples across different models}
    \label{tab:binary_choice_and_reason_examples}
    \resizebox{\textwidth}{!}{%
    \begin{tabular}{m{1.8cm} m{5cm} m{6.5cm} m{1.2cm} m{7.5cm}}
        \toprule
        \multicolumn{1}{c}{\textbf{Model}} & 
        \multicolumn{1}{c}{\textbf{Profile}} & 
        \multicolumn{1}{c}{\textbf{Retrieved Memory}} & 
        \multicolumn{1}{c}{\textbf{Choice}} & 
        \multicolumn{1}{c}{\textbf{Reason}} \\
        \midrule
        LLM-gpt4o & 
        Your name is Dorothy Roberts. You are a non-binary or other gender character, aged between 45 and 54, with a high income level, retired, with an associate degree, risk-neutral, and traveling for shopping. You are a character who is introverted, antagonistic, unconscientious, neurotic, and open to experience. & 
        Yesterday: route 1's travel time was 6, route 2's travel time was 38, and you chose route 2. Yesterday, you received a 0.04 RMB bonus, bringing your cumulative bonus to 7.28 RMB. Your historical travel experiences for each route over the past 47 days are as follows: route 1: Chosen 21 times, with an Experience Weighted Moving Average Travel Time of 33.35; route 2: Chosen 25 times, with an Experience Weighted Moving Average Travel Time of 31.83. &
        Route 1 & 
        Route 2 has a slightly lower Experience Weighted Moving Average Travel Time (31.83) compared to route 1 (33.35). However, yesterday's travel time for route 2 was significantly higher (38) compared to route 1 (6). Given the need for randomness and the recent poor performance of route 2, I will choose route 1 today. \\
        \midrule
        LLM-gpt35 & 
        Your name is Sandra Flores. You are a non-binary or other gender character, aged between 45 and 54, with a low income level, an employee, with an associate degree, risk-averse, and traveling for leisure. You are a character who is extroverted, agreeable, unconscientious, emotionally stable, and closed to experience. &
        Yesterday: route 1's travel time was 24, route 2's travel time was 20, and you chose route 1. Yesterday, you received a 0.32 RMB bonus, bringing your cumulative bonus to 7.80 RMB. Your historical travel experiences for each route over the past 33 days are as follows: route 1: Chosen 19 times, with an Experience Weighted Moving Average Travel Time of 28.23; route 2: Chosen 13 times, with an Experience Weighted Moving Average Travel Time of 25.91. &
        Route 2 &
        Based on the historical travel data, route 2 has a lower Experience Weighted Moving Average Travel Time compared to route 1. Additionally, since I am risk-averse and emotionally stable, I prefer to minimize uncertainty and stick with the option that has shown better performance so far. \\
        \midrule
        LLM-yi-medium & 
        Your name is Richard Lopez. You are a non-binary or other gender character, aged between 45 and 54, with a high income level, retired, with a doctorate, risk-neutral, and traveling for leisure. You are a character who is extroverted, agreeable, unconscientious, neurotic, and open to experience. &
        Yesterday: route 1's travel time was 28, route 2's travel time was 16, and you chose route 1. Yesterday, you received a 0.24 RMB bonus, bringing your cumulative bonus to 16.76 RMB. Your historical travel experiences for each route over the past 51 days are as follows: route 1: Chosen 25 times, with an Experience Weighted Moving Average Travel Time of 27.55; route 2: Chosen 25 times, with an Experience Weighted Moving Average Travel Time of 19.86. &
        Route 1 &
        Given the historical data and the current state of the game, route 2 offers a significantly shorter travel time on average, which would likely result in a higher bonus. However, considering the risk-neutral nature and the desire for randomness in decision-making, a strategic choice involves balancing the historical averages with the potential for a change in behavior among other participants. Given the extroverted and agreeable personality traits, there might be a subconscious inclination towards choosing the route that could potentially lead to a more harmonious outcome, even if it means sacrificing a slight advantage in travel time. Therefore, the choice is made with a slight preference for route 1, which has a slightly longer average travel time but could potentially lead to a more balanced distribution of traffic, thus maximizing the overall benefit for all participants. \\
        \midrule
        LLM-Llama3.1-70b &
        Your name is James Williams. You are a male character, aged between 45 and 54, with a middle income level, an employee, with a high school education, risk-averse, and traveling for business. You are a character who is introverted, antagonistic, conscientious, emotionally stable, and closed to experience. &
        Yesterday: route 1's travel time was 30, route 2's travel time was 14, and you chose route 1. Yesterday, you received a 0.2 RMB bonus, bringing your cumulative bonus to 18.84 RMB. Your historical travel experiences for each route over the past 61 days are as follows: route 1: Chosen 34 times, with an Experience Weighted Moving Average Travel Time of 26.65; route 2: Chosen 26 times, with an Experience Weighted Moving Average Travel Time of 21.32. &
        Route 2 &
        Considering my historical data, route 2 has a lower Experience Weighted Moving Average Travel Time (21.32) compared to route 1 (26.65). Additionally, yesterday's travel time for route 2 was significantly lower (14) than route 1 (30), which may indicate a trend. However, to incorporate randomness and avoid being too predictable, I will make a probabilistic decision. I will choose route 2 with a probability of 0.7 and route 1 with a probability of 0.3, but since route 2 has a clear advantage, I will choose it today. \\
        \midrule
        LLM-Llama3.1-8b &
        Your name is Carol King. You are a female character, aged between 25 and 34, with a middle income level, self-employed, with a bachelor's degree, risk-neutral, and traveling for education. You are a character who is introverted, agreeable, unconscientious, emotionally stable, and closed to experience. &
        Yesterday: route 1's travel time was 34, route 2's travel time was 10, and you chose route 1. Yesterday, you received a 0.12 RMB bonus, bringing your cumulative bonus to 20.68 RMB. Your historical travel experiences for each route over the past 69 days are as follows: route 1: Chosen 52 times, with an Experience Weighted Moving Average Travel Time of 24.14; route 2: Chosen 16 times, with an Experience Weighted Moving Average Travel Time of 23.86. &
        Route 1 &
        Given the historical data, route 1 has been chosen more frequently and has a slightly lower Experience Weighted Moving Average Travel Time compared to route 2. However, the difference is minimal, and I should consider the potential impact of others' choices on both routes. To maximize my benefit, I will choose the route with the lower travel time today, but I will also introduce an element of randomness to account for the uncertainty in others' decisions. \\
        \bottomrule
    \end{tabular}%
    }
\end{table}

\subsubsection{Descriptive statistics of experiment results}

Table~\ref{tab:tt_statistics} and Figure~\ref{fig:travel_time_evolutions} shows the travel time on "route 1" in each scenario for different LLMTravelers of the experimental data. Similar to laboratory findings in \cite{qi2023investigating}, the mean travel times for all LLMTravelers closely approximate the DUE points. Routes with higher flow sensitivity display greater variance in travel time. For instance, the standard deviation of travel time for "route 2" consistently exceeds that of "route 1", reflecting "route 2"'s greater sensitivity to flow changes.

In comparing results across models, the "\((\text{Mean} - \text{DUE}) / \text{DUE}\)" column in Table~\ref{tab:tt_statistics} measures the relative difference between each model's mean travel time and the DUE point, providing an indicator of alignment with equilibrium conditions. In the laboratory results, values for "\((\text{Mean} - \text{DUE}) / \text{DUE}\)" remain within \(\pm 2\%\), indicating close alignment with the DUE. By contrast, LLMTravelers tend to exhibit slightly larger deviations from the DUE, with most values remaining within \(\pm 10\%\). Additionally, the laboratory results consistently yields the smallest standard deviation in comparison to the LLMTravelers.

A distinctive case is presented by the LLM-gpt4o agent. While its "\((\text{Mean} - \text{DUE}) / \text{DUE}\)" ratio remains relatively low, indicating its mean travel time is near the DUE, its standard deviation is the highest among all scenarios. This suggests greater daily deviations from the DUE. Figure~\ref{fig:travel_time_evolutions} (b) further illustrates the fluctuation pattern of LLM-gpt4o, showing that daily travel times periodically evolve around the DUE. This behavior may result from LLM-gpt4o agents adhering to a fixed set of strategies, leading to cyclical patterns over time.

This pattern may not be ideal within this experimental context, as the LLMTraveler's prompt include profiles do not introduce sufficient variability. As a result, agents with different profiles exhibit minimal behavioral differences under similar experiences. In particular, LLMTravelers based on "GPT-4o-2024-05-13" \cite{achiam2023gpt} tend to select the same route despite profile differences, which deviates from realistic behavior patterns. This outcome indicates that larger, more complex pretrained models do not necessarily enhance the realism of simulated route-switching behavior. However, as shown in Figure~\ref{fig:travel_time_evolutions}, other LLM-based agents effectively reproduce stochastic fluctuations around the equilibrium point, which is similar to previous laboratory data \cite{selten2007commuters, meneguzzer2013day, qi2023investigating, dixit2014equilibrium}.

\begin{table}[h]
    \centering
    \small
    \caption{Descriptive statistics of travel time on different routes in each scenario for different LLMTraveler}
    \label{tab:tt_statistics}
    \resizebox{\textwidth}{!}{ 
    \begin{tabular}{cccccccccc}
        \toprule
        \textbf{Scenario} & \textbf{Model} & \multicolumn{4}{c}{\textbf{Route 1}} & \multicolumn{4}{c}{\textbf{Route 2}} \\
        \cmidrule(lr){3-6} \cmidrule(lr){7-10}
         &  & \textbf{DUE\textsuperscript{a}} & \textbf{Mean\textsuperscript{b}} & \textbf{(Mean-DUE)/DUE\textsuperscript{c}} & \textbf{Std\textsuperscript{d}} & \textbf{DUE\textsuperscript{a}} & \textbf{Mean\textsuperscript{b}} & \textbf{(Mean-DUE)/DUE\textsuperscript{c}} & \textbf{Std\textsuperscript{d}} \\
        \midrule
          & Lab & 22.00 & 22.08 & 0.36\% & 3.55 & 22.00 & 21.92 & -0.36\% & 3.55 \\
          & LLM-gpt4o & 22.00 & 22.29 & 1.33\% & 13.23 & 22.00 & 21.71 & -1.33\% & 13.23 \\
        1 & LLM-gpt35 & 22.00 & 23.96 & 8.91\% & 5.26 & 22.00 & 20.04 & -8.91\% & 5.26 \\
          & LLM-llama-3.1-70b & 22.00 & 23.66 & 7.55\% & 4.66 & 22.00 & 20.34 & -7.55\% & 4.66 \\
          & LLM-yi-medium & 22.00 & 26.73 & 21.48\% & 4.97 & 22.00 & 17.27 & -21.48\% & 4.97 \\
          & LLM-llama-3.1-8b & 22.00 & 23.07 & 4.88\% & 6.18 & 22.00 & 20.93 & -4.88\% & 6.18 \\
        \midrule
          & Lab & 54.00 & 53.59 & -0.76\% & 6.09 & 54.00 & 54.61 & 1.13\% & 9.14 \\
          & LLM-gpt4o & 54.00 & 53.08 & -1.70\% & 20.36 & 54.00 & 55.38 & 2.56\% & 30.54 \\
        2 & LLM-gpt35 & 54.00 & 55.85 & 3.43\% & 8.71 & 54.00 & 51.22 & -5.15\% & 13.07 \\
          & LLM-llama-3.1-70b & 54.00 & 51.65 & -4.35\% & 14.57 & 54.00 & 57.52 & 6.52\% & 21.85 \\
          & LLM-yi-medium & 54.00 & 57.35 & 6.20\% & 8.50 & 54.00 & 48.98 & -9.30\% & 12.75 \\
          & LLM-llama-3.1-8b & 54.00 & 55.23 & 2.27\% & 10.67 & 54.00 & 52.16 & -3.41\% & 16.00 \\
        \midrule
          & Lab & 27.00 & 26.83 & -0.63\% & 3.42 & 27.00 & 27.25 & 0.93\% & 5.13 \\
          & LLM-gpt4o & 27.00 & 26.17 & -3.06\% & 10.07 & 27.00 & 28.24 & 4.59\% & 15.10 \\
        3 & LLM-gpt35 & 27.00 & 28.30 & 4.81\% & 4.64 & 27.00 & 25.05 & -7.22\% & 6.96 \\
          & LLM-llama-3.1-70b & 27.00 & 25.07 & -7.16\% & 6.95 & 27.00 & 29.90 & 10.74\% & 10.42 \\
          & LLM-yi-medium & 27.00 & 28.42 & 5.26\% & 4.38 & 27.00 & 24.87 & -7.89\% & 6.56 \\
          & LLM-llama-3.1-8b & 27.00 & 27.88 & 3.26\% & 5.70 & 27.00 & 25.68 & -4.89\% & 8.55 \\
        \midrule
          & Lab & 55.20 & 54.35 & -1.54\% & 7.52 & 55.20 & 56.48 & 2.32\% & 11.28 \\
          & LLM-gpt4o & 55.20 & 54.20 & -1.81\% & 21.29 & 55.20 & 56.70 & 2.72\% & 31.93 \\
        4 & LLM-gpt35 & 55.20 & 57.16 & 3.55\% & 8.72 & 55.20 & 52.26 & -5.33\% & 13.08 \\
          & LLM-llama-3.1-70b & 55.20 & 54.25 & -1.71\% & 13.91 & 55.20 & 56.62 & 2.57\% & 20.87 \\
          & LLM-yi-medium & 55.20 & 59.01 & 6.91\% & 8.92 & 55.20 & 49.48 & -10.36\% & 13.38 \\
          & LLM-llama-3.1-8b & 55.20 & 56.68 & 2.68\% & 11.09 & 55.20 & 52.98 & -4.02\% & 16.64 \\
        \midrule
          & Lab & 27.60 & 27.21 & -1.41\% & 3.90 & 27.60 & 28.18 & 2.10\% & 5.84 \\
          & LLM-gpt4o & 27.60 & 26.66 & -3.41\% & 11.00 & 27.60 & 29.01 & 5.11\% & 16.49 \\
        5 & LLM-gpt35 & 27.60 & 28.66 & 3.84\% & 4.35 & 27.60 & 26.01 & -5.76\% & 6.53 \\
          & LLM-llama-3.1-70b & 27.60 & 26.41 & -4.32\% & 6.83 & 27.60 & 29.39 & 6.49\% & 10.25 \\
          & LLM-yi-medium & 27.60 & 29.57 & 7.13\% & 4.69 & 27.60 & 24.65 & -10.69\% & 7.03 \\
          & LLM-llama-3.1-8b & 27.60 & 28.39 & 2.85\% & 5.34 & 27.60 & 26.42 & -4.28\% & 8.01 \\
        \bottomrule
    \end{tabular}
    }
    \vspace{0.2em}
    \textsuperscript{a} Travel time under DUE. \\
    \textsuperscript{b} The average travel time observed in the experiment. \\
    \textsuperscript{c} The percentage difference between the Mean and DUE travel time. \\
    \textsuperscript{d} The standard deviation of travel time in the experiment.
\end{table}

\begin{figure}
  \centering
  \includegraphics[width=0.98\textwidth]{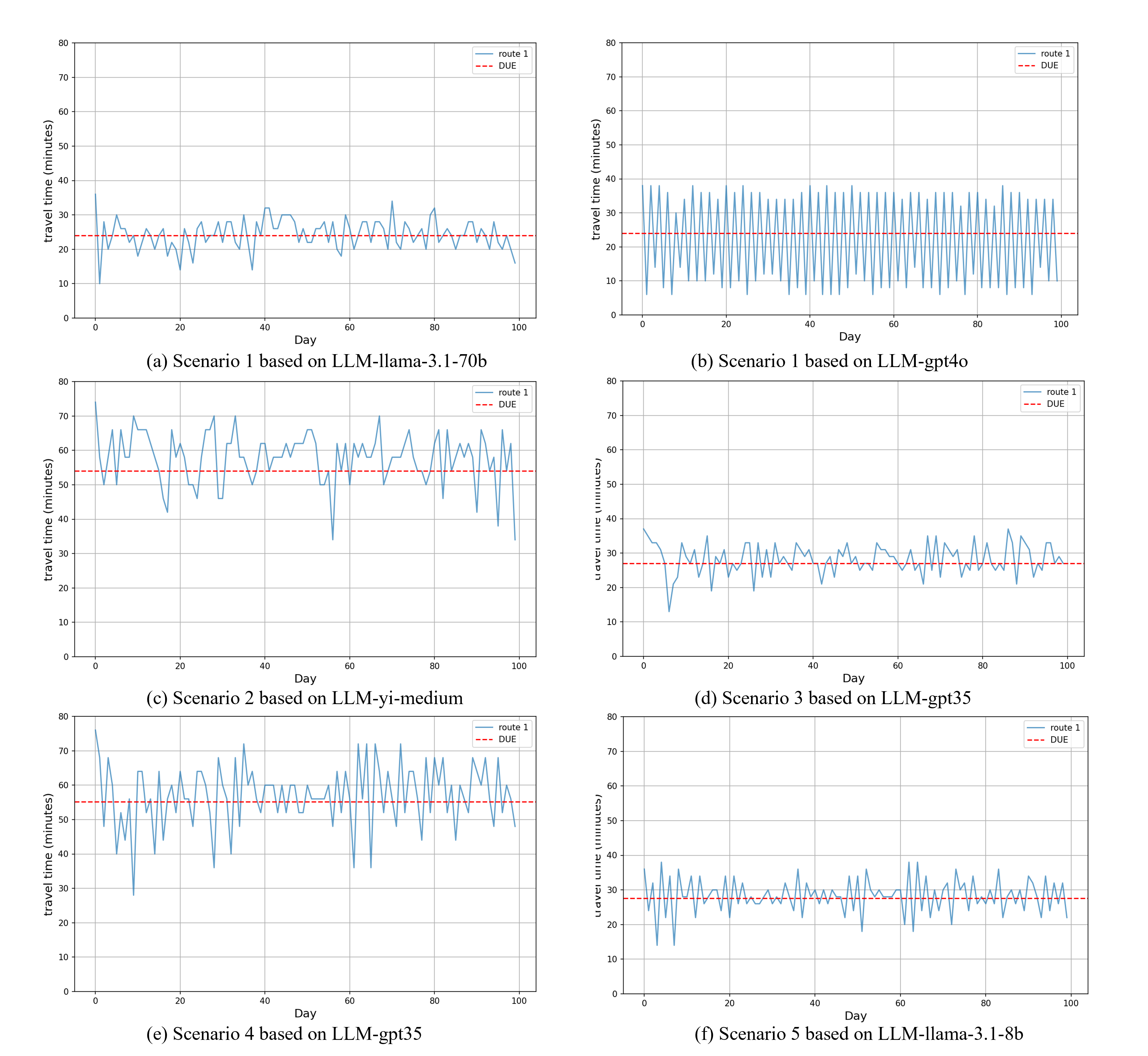}
  \caption{Travel time evolution examples in all scenarios in the experiment}
  \label{fig:travel_time_evolutions}
\end{figure}

\begin{figure}
  \centering
  \includegraphics[width=0.98\textwidth]{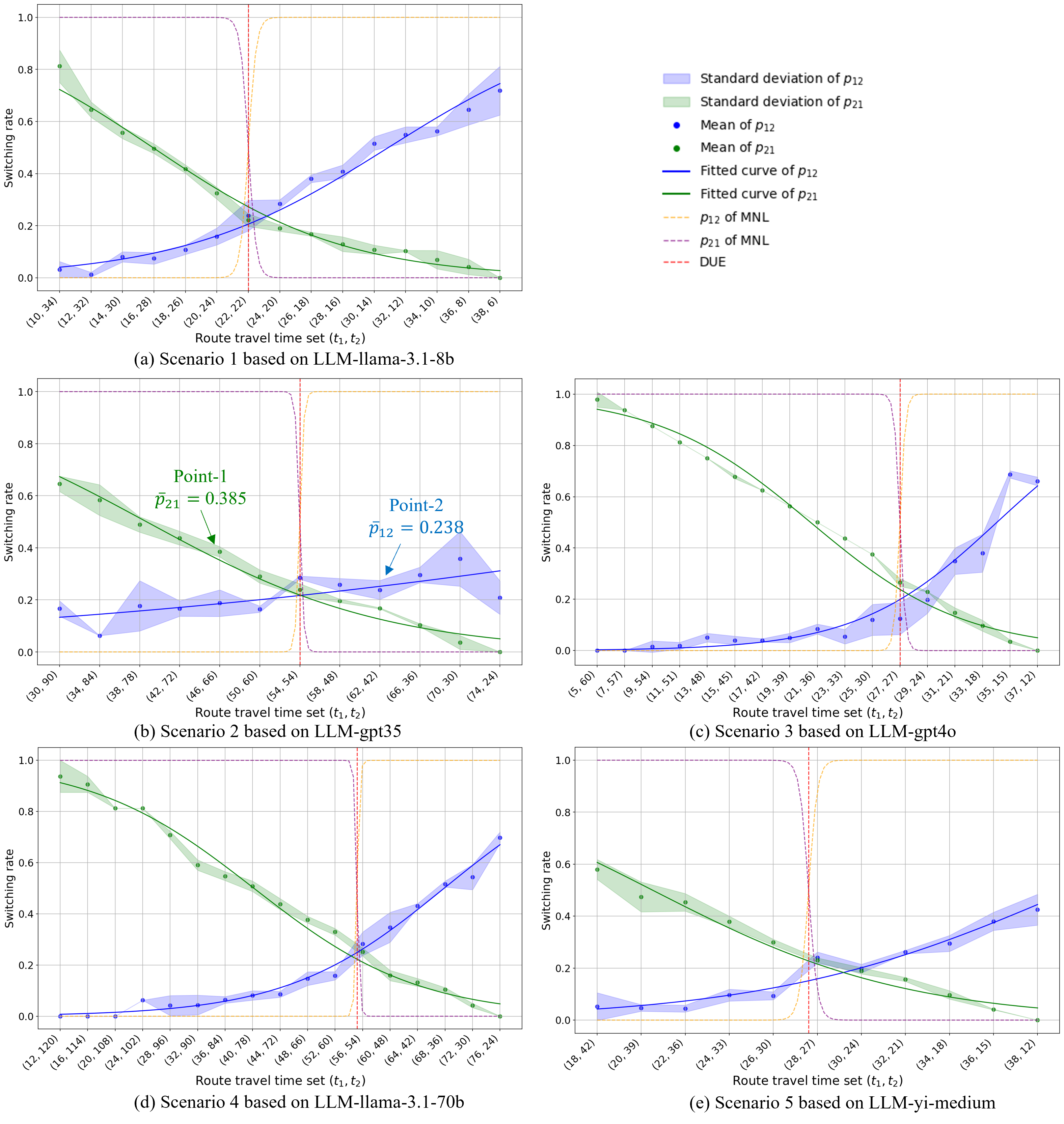}
  \caption{Observed switching rates examples of all scenarios}
  \label{fig:switching_rate}
\end{figure}

\subsubsection{Route switching behavior analysis}

This subsection presents an analysis of observed switching rates to identify potential similarities and differences between laboratory observations, established theories, and the proposed LLMTraveler ageny. The focus is on examining the relationship between average switching rates and different cost combinations. Specifically, three route-switching behavior patterns observed in laboratory data \cite{qi2023investigating} are compared to those exhibited by LLMTravelers. Following the approach in \cite{qi2023investigating}, the logistic regression is applied to model the binary decision-making process in the scenarios. The switching rate is fitted using maximum likelihood estimation across all experimental data:
\begin{equation}
    p_{ij} = \frac{1}{1 + e^{-(\theta_0 + \theta_1 (t_i - t_j))}}
\end{equation}
where $\theta_0$ and $\theta_1$ are the parameters to be estimated, and $t_i$ and $t_j$ represent the travel times on route $i$ and route $j$, respectively.

This model enables a statistical interpretation of choice behavior by incorporating the effects of route cost differences on decision-making. In contrast to the simple averaging method, this approach adjusts for varying frequencies of different cost combinations, assigning appropriate weights based on occurrence. Figure~\ref{fig:switching_rate} shows the fitted logistic curve, with estimated parameter values detailed in Table~\ref{tab:logit_func_statistics}. In the laboratory results, all \( p \)-values are less than 0.01, indicating that all variables are statistically significant. However, in the LLMTraveler-based experiments, not all variables achieve significance; some models have \( p \)-values greater than 0.05, particularly LLM-gpt35 and LLM-yi-medium. Most other models, however, exhibit significant \( p \)-values.  Despite these differences, the logistic function effectively captures the overall trend of the experimental data, as shown by the fitted curves in Figure~\ref{fig:switching_rate}. This holds true even for the fitted curves of LLM-gpt35 in Scenario 2 (Figure~\ref{fig:switching_rate}(b)) and LLM-yi-medium in Scenario 4 (Figure~\ref{fig:switching_rate}(e)). Additionally, the fitted parameters vary among different LLMTravelers, demonstrating that each agent has a different sensitivity to the same route conditions.

\begin{table}[h]
    \centering
    \caption{Regression results for all scenarios with different LLM models}
    \label{tab:logit_func_statistics}
    \resizebox{\textwidth}{!}{ 
    \begin{tabular}{cccccccccccccc}
        \toprule
        \textbf{Scenario} & \textbf{Model} & \multicolumn{4}{c}{$p_{12}$} & \multicolumn{4}{c}{$p_{21}$} \\
        \cmidrule(lr){3-6} \cmidrule(lr){7-10}
         &  & \textbf{$\theta_0$} & \textbf{$\theta_1$} & \textbf{p value of $\theta_0$} & \textbf{p value of $\theta_1$} & \textbf{$\theta_0$} & \textbf{$\theta_1$} & \textbf{p value of $\theta_0$} & \textbf{p value of $\theta_1$} \\
        \midrule
          & Lab & -0.7730 & 0.0324 & <0.001 & <0.001 & -0.7410 & 0.0347 & <0.001 & <0.001 \\
          & LLM-gpt4o & -0.9726 & 0.1095 & 0.1124 & 0.0008 & -0.3682 & 0.0879 & 0.4405 & 0.0004 \\
        1 & LLM-gpt35 & -0.8976 & 0.0105 & 0.0210 & 0.6353 & -0.4268 & 0.0690 & 0.2619 & 0.0118 \\
          & LLM-llama-3.1-70b & -1.0153 & 0.0502 & 0.0161 & 0.0521 & -0.7314 & 0.0602 & 0.0566 & 0.0269 \\
          & LLM-yi-medium & -1.6720 & 0.0442 & 0.0018 & 0.1072 & -0.7205 & 0.0760 & 0.0779 & 0.0089 \\
          & LLM-llama-3.1-8b & -1.3487 & 0.0758 & 0.0042 & 0.0049 & -0.9815 & 0.0809 & 0.0163 & 0.0053 \\
        \midrule
          & Lab & -1.5640 & 0.0175 & <0.001 & <0.001 & -0.6920 & 0.0290 & <0.001 & <0.001 \\
          & LLM-gpt4o & -1.6332 & 0.0403 & 0.0017 & 0.0098 & -1.1851 & 0.0362 & 0.0129 & 0.0003 \\
        2 & LLM-gpt35 & -1.2841 & 0.0098 & 0.0018 & 0.4171 & -1.2739 & 0.0333 & 0.0086 & 0.0157 \\
          & LLM-llama-3.1-70b & -0.9464 & 0.0335 & 0.0203 & 0.0054 & -1.2816 & 0.0300 & 0.0066 & 0.0019 \\
          & LLM-yi-medium & -1.6751 & 0.0159 & 0.0006 & 0.2922 & -1.3990 & 0.0358 & 0.0064 & 0.0227 \\
          & LLM-llama-3.1-8b & -1.3918 & 0.0241 & 0.0020 & 0.0745 & -1.4517 & 0.0379 & 0.0060 & 0.0068 \\
        \midrule
          & Lab & -1.4350 & 0.0334 & <0.001 & <0.001 & -0.3990 & 0.0318 & <0.001 & <0.001 \\
          & LLM-gpt4o & -1.3917 & 0.0790 & 0.0049 & 0.0059 & -1.1593 & 0.0714 & 0.0184 & 0.0003 \\
        3 & LLM-gpt35 & -1.3447 & 0.0140 & 0.0019 & 0.5927 & -1.2420 & 0.0550 & 0.0078 & 0.0458 \\
          & LLM-llama-3.1-70b & -0.6674 & 0.0609 & 0.0901 & 0.0080 & -1.2160 & 0.0542 & 0.0101 & 0.0041 \\
          & LLM-yi-medium & -1.8581 & 0.0539 & 0.0010 & 0.1163 & -1.4209 & 0.0746 & 0.0066 & 0.0151 \\
          & LLM-llama-3.1-8b & -1.5040 & 0.0541 & 0.0018 & 0.0657 & -1.3970 & 0.0774 & 0.0075 & 0.0077 \\
        \midrule
          & Lab & -1.1660 & 0.0127 & <0.001 & <0.001 & -0.1490 & 0.0078 & <0.01 & <0.01 \\
          & LLM-gpt4o & -1.7908 & 0.0471 & 0.0047 & 0.0086 & -1.0800 & 0.0361 & 0.0296 & 0.0003 \\
        4 & LLM-gpt35 & -1.2693 & 0.0121 & 0.0034 & 0.3636 & -1.1603 & 0.0315 & 0.0119 & 0.0336 \\
          & LLM-llama-3.1-70b & -1.1032 & 0.0349 & 0.0097 & 0.0053 & -1.2503 & 0.0333 & 0.0082 & 0.0008 \\
          & LLM-yi-medium & -1.6964 & 0.0217 & 0.0016 & 0.1851 & -1.2779 & 0.0284 & 0.0068 & 0.0571 \\
          & LLM-llama-3.1-8b & -1.3157 & 0.0229 & 0.0027 & 0.0740 & -1.3571 & 0.0380 & 0.0077 & 0.0058 \\
        \midrule
          & Lab & -1.2140 & 0.0216 & <0.001 & <0.001 & -0.2620 & 0.0190 & <0.001 & <0.001 \\
          & LLM-gpt4o & -1.5448 & 0.0911 & 0.0104 & 0.0056 & -1.1060 & 0.0729 & 0.0381 & 0.0006 \\
        5 & LLM-gpt35 & -1.2013 & 0.0199 & 0.0037 & 0.4313 & -1.1287 & 0.0652 & 0.0138 & 0.0258 \\
          & LLM-llama-3.1-70b & -0.8910 & 0.0693 & 0.0313 & 0.0065 & -1.2278 & 0.0578 & 0.0083 & 0.0041 \\
          & LLM-yi-medium & -1.7247 & 0.0578 & 0.0019 & 0.0856 & -1.2217 & 0.0690 & 0.0105 & 0.0310 \\
          & LLM-llama-3.1-8b & -1.4870 & 0.0557 & 0.0015 & 0.0412 & -1.3283 & 0.0751 & 0.0083 & 0.0036 \\
        \bottomrule
    \end{tabular}
    }
\end{table}

\subsubsection{Observed route-switching behavior patterns}

This subsection evaluates whether the LLMTraveler exhibits consistency with the three route-switching behavior patterns identified in the laboratory data \citep{qi2023investigating}.

\textbf{First pattern}. 
The switching rate increases with the cost difference between the last-chosen route and its alternative. Even when the last-chosen route has a lower cost than the alternative, the switching rate remains positive. The PRC model fails to capture this pattern, as it assumes travelers will only switch to route with lower cost \citep{zhang2001equivalence}. By contrast, the MNL model accounts for this behavior, as it allows travelers to switch even when their previous choice is better. This pattern is replicated across all LLMTravelers, as shown in Figure~\ref{fig:switching_rate}. Moving from left to right along the x-axis, where the cost of "route 1" increases and that of "route 2" decreases, the switching rate from "route 1" to "route 2" rises, while the switching rate from "route 2" to "route 1" declines. Additionally, as shown in Figure~\ref{fig:switching_rate}(b), even when the cost of "route 1" is 30 minutes and "route 2" is 90 minutes (three times greater), some LLMTravelers still switch from "route 1" to "route 2."

\textbf{Second pattern}. 
In asymmetric networks, the average switching rates at the DUE point for both routes are between 0 and 0.5, but with significant differences between \( p_{12} \) and \( p_{21} \). In the traditional PRC and MNL models, the DUE switching rates are assumed to be 0 and 0.5, respectively, which does not align with these laboratory observations. For the LLMTravelers, as shown in Figure~\ref{fig:switching_rate} (b–e), the average switching rates at the DUE point also fall between 0 and 0.5. However, unlike the laboratory data, the values of \( p_{12} \) and \( p_{21} \) are close to each other, especially in Figure~\ref{fig:switching_rate}(d) and Figure~\ref{fig:switching_rate}(e), where the observed \( p_{12} \) and \( p_{21} \) values at the DUE point are nearly identical.

\textbf{Third pattern}. 
The average switching rate is influenced not only by the cost difference but also by the characteristics of the last-chosen route. For instance, in Scenario 2 of laboratory data, when the cost of "route 2" is 20 minutes higher than "route 1", the average switching rate from "route 2" to "route 1" is 0.481. In contrast, when the cost of "route 1" is 20 minutes higher than "route 2," the switching rate from "route 1" to "route 2" drops to 0.267 \citep{qi2023investigating}. This pattern is also observed among LLMTravelers, as shown in Figure~\ref{fig:switching_rate}(b). "Point-1" indicates that when the cost of "route 2" is 20 minutes higher than "route 1", the switching rate from "route 2" to "route 1" is 0.385. In contrast, when the cost of "route 1" is 20 minutes higher than "route 2," the switching rate from "route 1" to "route 2" drops to 0.238.

In summary,the LLMTravelers' route-switching behavior aligns with most observed patterns from laboratory data, except for the absence of a distinct difference between \( p_{12} \) and \( p_{21} \) at the DUE point in asymmetric networks. These patterns, which are not fully captured by traditional models, demonstrate that LLMTravelers choose routes in a logit-like manner, showing "inertia" by favoring previously chosen routes \citep{qi2023investigating}. Additionally, choices are influenced not only by past route costs but also by inherent route characteristics, indicating that travelers treat routes with different attributes differently.

\section{Evaluation of UE choices in a multi-OD pair network}
\label{sec:eval-multiOD}

This section explores the application of the proposed LLMTraveler to model the learning behaviors of selfish agents in the multi-OD pair congestion game. 

\subsection{Experiment Settings}

Figure~\ref{fig:OW} illustrates the OW road network \citep{de2024modelling}, which connects two residential areas (1 and 2) with two large shopping centers (12 and 13). The figure also displays the free-flow travel times between these nodes, measured in minutes. Notably, all links in the network are two-way. The performance function for each link is the free-flow travel time plus 0.02 minutes for each vehicle per hour of flow. Table~\ref{tab:OW_od_demand} presents the OD demand for the OW network. For example, the first row indicates that a total of 600 travelers depart from node 1 to node 12, making the entire network accommodate 1,700 travelers. For each OD pair, the \( k \)-shortest path algorithm \citep{yen1970algorithm} is used to determine their route choice sets. In the experiments, \( k=5 \), meaning each OD pair has 5 predefined routes to choose from.

\begin{figure}
  \centering
  \includegraphics[width=0.45\textwidth]{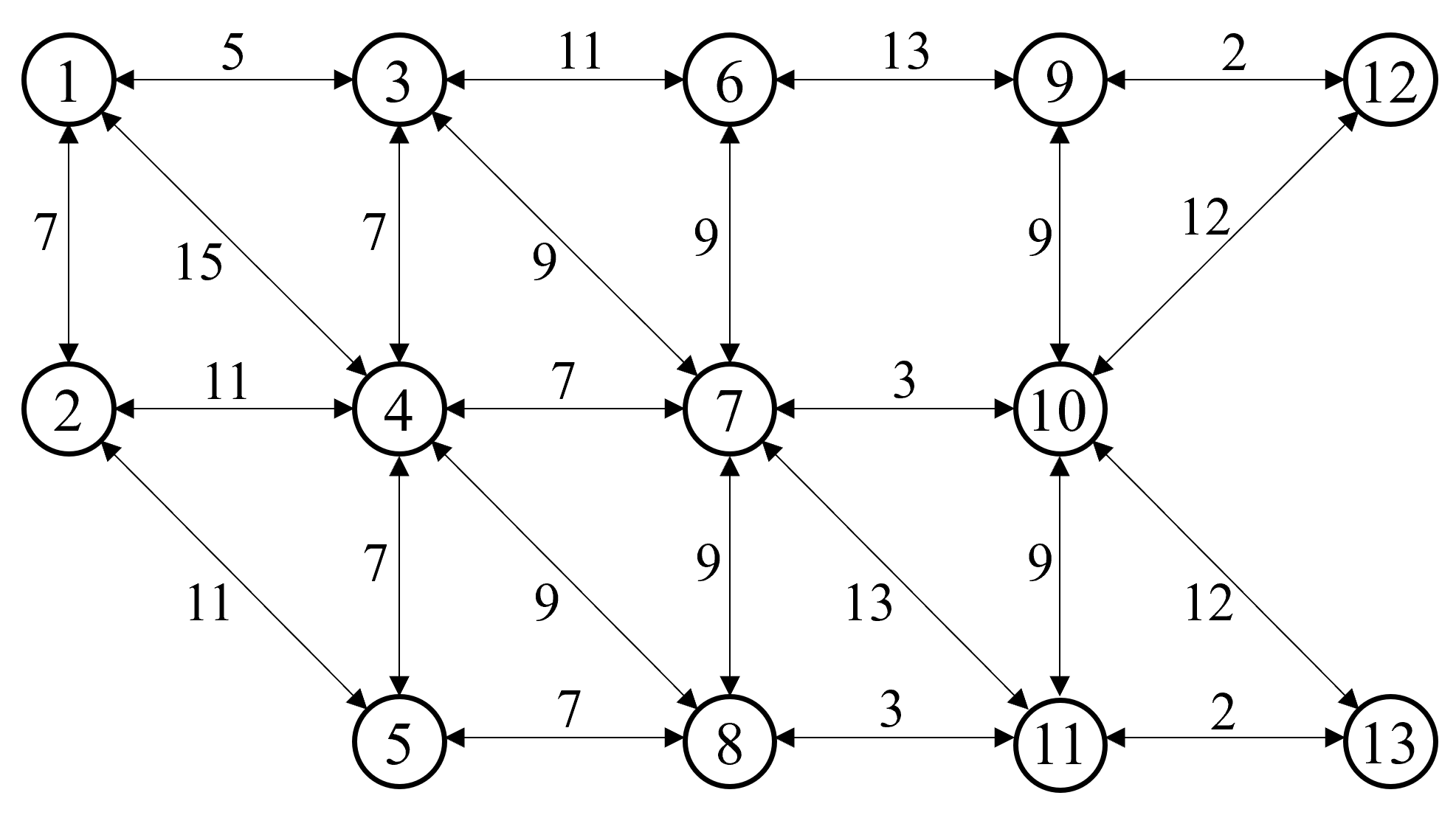}
  \caption{OW network}
  \label{fig:OW}
\end{figure}

\begin{table}[h]
    \centering
    \small 
    \caption{OD demand of OW network}
    \label{tab:OW_od_demand}
    \begin{tabular}{cccc}
        \toprule
        \textbf{Origin} & \textbf{Destination} & \textbf{Demand} & \textbf{Number of agents} \\
        \midrule
        1 & 12 & 600 & 30 \\
        1 & 13 & 400 & 20 \\
        2 & 12 & 300 & 15 \\
        2 & 13 & 400 & 20 \\
        \bottomrule
    \end{tabular}
\end{table}

To simplify the simulation process, a LLMTraveler is used to represent \( n \) (e.g., 10, 20, 50) travelers, meaning the decisions of one LLMTraveler reflect those of \( n \) travelers. Different values of \( n \) were tested, and as shown in Figure~\ref{fig:diff_n}, varying \( n \) does not affect the convergence of the system towards UE. Therefore, for subsequent experiments, \( n=20 \) is used, which balances experimental cost and time. Consequently, the number of agents corresponding to each OD pair is shown in the last column of Table~\ref{tab:OW_od_demand}.

The prompt template used in this experiment is shown in Figure~\ref{fig:prompt}. Considering that most studies view travelers as selfish, the profile includes the phrase "you are selfish." The LLM temperature was set to 0.5 by default in this experiment, while the smoothing factor \(\omega\) used for calculating EWMATT was set to 0.2. Additionally, all LLMTravelers are simulated for 100 days per run, with a total of three runs conducted to ensure robust results.

\begin{figure}
  \centering
  \includegraphics[width=0.8\textwidth]{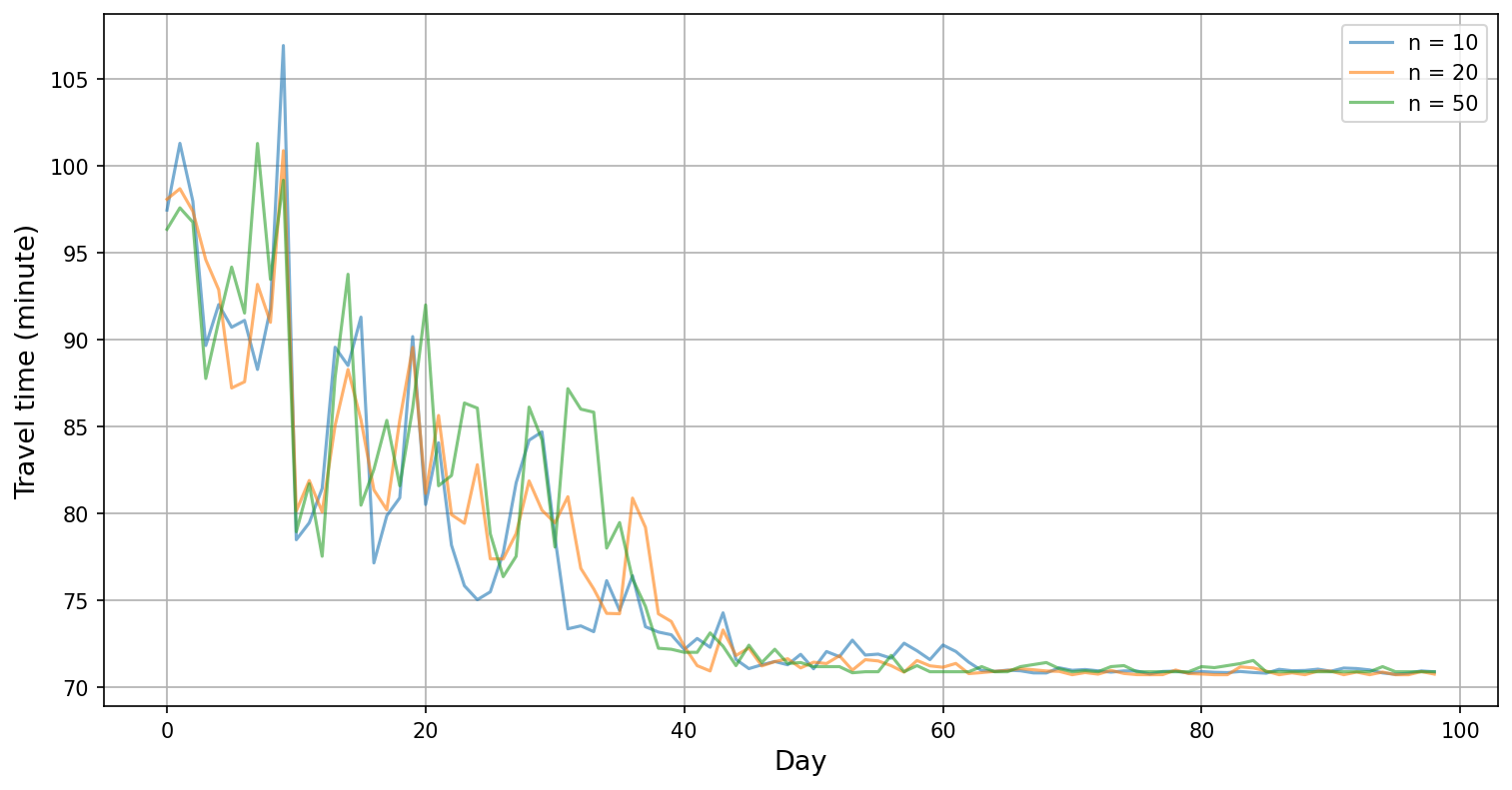}
  \caption{Travel time over days with different $n$}
  \label{fig:diff_n}
\end{figure}

\subsection{Evaluation Metrics}
In this section, travel time and Day Switching Rate (DSR) are used for evaluation.

\textbf{Travel time}. 
This metric calculates the average travel time of all travelers collected on the same day.

\textbf{DSR}. 
This metric builds upon the general idea of the Switching Rate. To measure the switching behavior of travelers who switch routes from day \( t \) to day \( t+1 \), the DSR \( R^t \) is defined. Let \( N \) be the total number of travelers, and \( n_{\text{switch}}^t \) be the number of travelers who switched their route from day \( t \) to day \( t+1 \). The DSR \( R^t \) is defined as:
\begin{equation}
    R^t = \frac{n_{\text{switch}}^t}{N}
    \label{eq:dtdsr}
\end{equation}

\subsection{Compared Method}

In addition to the LLMTraveler and MNL model discussed in Subsection~\ref{subsec:eval_compared_method}, the performance of the proposed method is also compared with the following RL-based approach:

\textbf{RL-Based Method}. 
In this approach, travelers are treated as agents to model the learning behaviors of selfish agents in congestion games \citep{ramos2018analysing, mao2018reinforcement, zhou2020reinforcement, shou2022multi}. This method adopts commonly used settings from prior RL-based congestion game studies, where agents rely on local observations (e.g., origin and destination) to select a route from the available route set each day as their action. The environment provides feedback (e.g., negative travel time), which serves as the reward. Through trial-and-error interactions, the agents accumulate experience and adapt their strategies over time, aiming to minimize individual travel time. The agents are trained using the Independent Proximal Policy Optimization (IPPO) algorithm with parameter sharing to enhance learning efficiency \citep{yu2022surprising}.

\subsection{Experiment Results}

\subsubsection{Aggregate-level behavior of LLMTravelers}

Figure~\ref{fig:all_llms_tt_and_dsr}(a) shows that all LLMTravelers converge toward a smaller travel time, indicating a move towards UE, after experiencing initial fluctuations over several days. Although the travel time becomes smaller over time, fluctuations persist, which aligns with findings from laboratory experiments \citep{iida1992experimental, meneguzzer2013day}. This highlights an advantage of the proposed method compared to some traditional traffic assignment models that assume all travelers are selfish and cannot model these fluctuations. Figure~\ref{fig:all_llms_tt_and_dsr}(b) shows the changes in the DSR over time. Although the specific DSR values vary across different LLMTravelers, the overall trend is consistent: an initial increase followed by a decrease, eventually stabilizing. By day 100, all models except LLM-yi-medium exhibit a DSR below 0.2, with LLM-gpt4o approaching zero. The aggregate-level route choice behavior of LLMTravelers can be summarized into three stages:

\begin{figure}[h]
    \centering
    \includegraphics[width=0.8\textwidth]{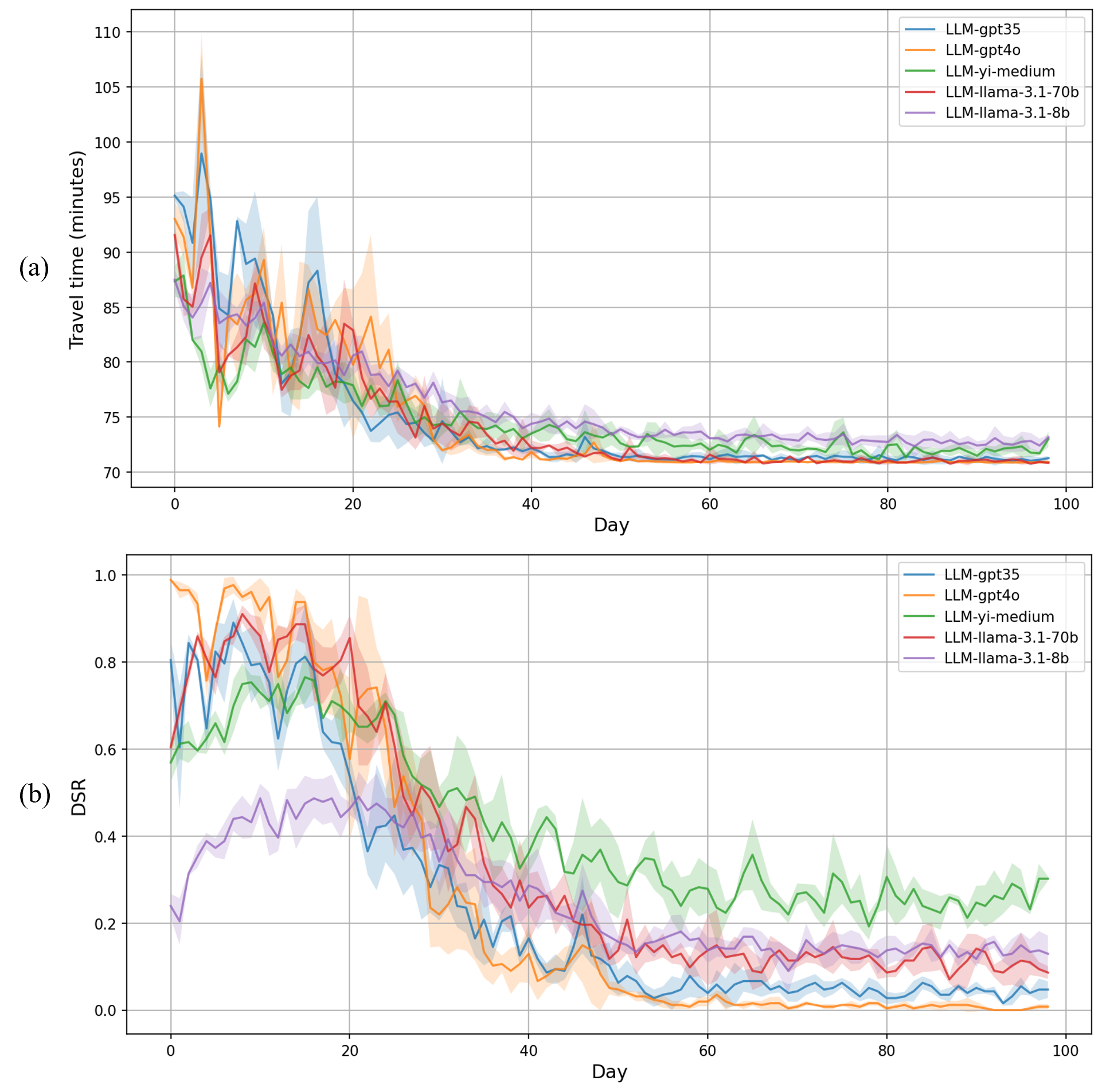}
    \caption{Travel time and DSR of all LLMTravelers over days}
    \label{fig:all_llms_tt_and_dsr}
\end{figure}

\textbf{Stage 1: Exploration (day 1 to 20)}.
During this initial phase, all LLMTravelers lack experience and knowledge about the available routes, leading to mostly random or exploratory route choices. This stage is characterized by attempts to explore routes with little to no prior experience. For instance, as shown in the first row of Table~\ref{tab:ow_route_choice_reason}, the LLMTraveler's reasoning is described as "any route can be chosen." Similarly, the second row illustrates an exploration-driven decision, where the LLMTraveler chooses to "explore route 2 despite its higher EWMATT," seeking to gather additional data and potentially uncover a new optimal route.

\textbf{Stage 2: Exploration and exploitation (day 21 to 60)}.
In this intermediate phase, LLMTravelers balance exploration and exploitation. As their experience grows, they begin to leverage knowledge of previously tested routes while still testing less frequently used routes with higher EWMATT values. This stage reflects a strategic approach where agents aim to refine their understanding of optimal routes by combining exploratory actions with exploitation of known information. Examples of both exploration and exploitation behaviors can be observed in the second and third rows of Table~\ref{tab:ow_route_choice_reason}.

\textbf{Stage 3: Exploitation (after day 60)}.
By this stage, most routes have been tried multiple times, and LLMTravelers have developed a reliable understanding of the best routes based on EWMATT. Consequently, the majority of choices are exploitative, although there remains a small probability of exploration, accounting for the slight fluctuations in the average travel time.

Figure~\ref{fig:avg_sr_first_last_n_days} further illustrates how the number of parameters of LLM models impacts the performance of LLMTravelers. Figure~\ref{fig:avg_sr_first_last_n_days}(a) reveals that the average DSR during the first 20 days increases with model size. On the other hand, Figure~\ref{fig:avg_sr_first_last_n_days}(b) shows that the average DSR during the last 20 days generally decreases with model size, except for the LLM-yi-medium model, which achieves a relatively higher DSR in this period. The LLM models in Figure~\ref{fig:avg_sr_first_last_n_days} are arranged in increasing order of parameter size, from 7 billion to over 175 billion. This may be because larger models, with their capacity to fit smaller errors, align more closely with human preferences through reinforcement learning from human feedback (RLHF). Such alignment can result in "overconfidence"~\citep{achiam2023gpt, kadavath2022language, leng2024taming}, reducing flexibility and limiting exploration. This behavior likely contributes to the observed decrease in DSR during the later stages. In contrast, smaller models maintain greater randomness, which may explain their relatively higher DSR during the final period.

\begin{figure}[h]
    \centering
    \includegraphics[width=0.8\textwidth]{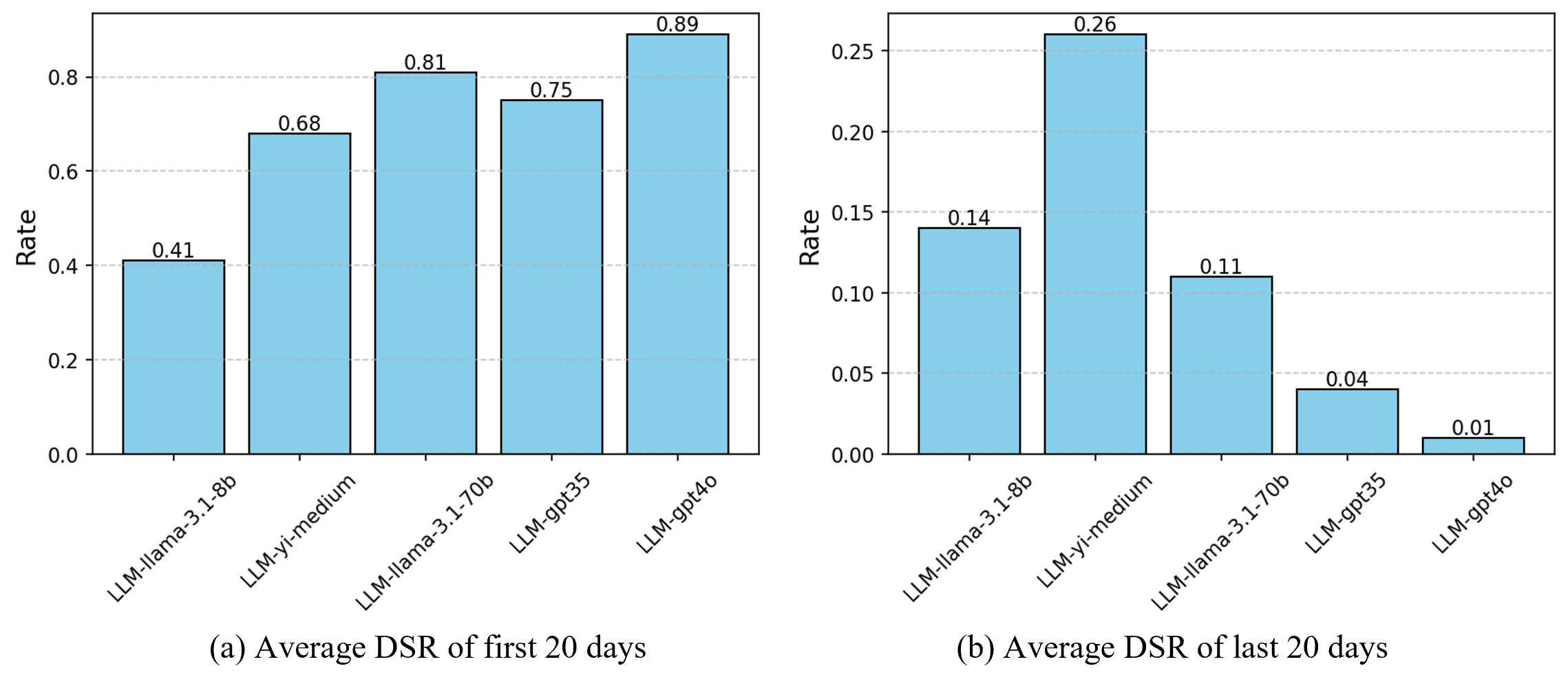}
    \caption{Average DSR of first 20 days and last 20 days}
    \label{fig:avg_sr_first_last_n_days}
\end{figure}

\begin{table}[h]
    \centering
    \small 
    \caption{Examples of LLMTraveler's route choice and reason}
    \label{tab:ow_route_choice_reason}
    \begin{tabular}{>{\centering\arraybackslash}p{1cm} >{\centering\arraybackslash}p{6cm} >{\centering\arraybackslash}p{1cm} >{\centering\arraybackslash}p{6cm}}
        \toprule
        \textbf{Type} & \textbf{Retrieved Memory} & \textbf{Choice} & \textbf{Reason} \\
        \midrule
        Random & 
        \begin{tabular}[c]{@{}c@{}}
            Route 0: Chosen 1 time, EWMATT = 50 \\
            Route 1: Chosen 1 time, EWMATT = 50 \\
            Route 2: Chosen 1 time, EWMATT = 50 \\
            Route 3: Chosen 1 time, EWMATT = 50 \\
            Route 4: Chosen 1 time, EWMATT = 50
        \end{tabular} & 
        Route 0 & 
        \begin{tabular}[c]{@{}c@{}}
            All routes have been chosen once and have \\
            equal EWMATT of 50. Since all routes are \\
            equally optimal based on historical data, \\
            any route can be chosen.
        \end{tabular} \\
        \midrule
        Explore & 
        \begin{tabular}[c]{@{}c@{}}
            Route 0: Chosen 18 times, EWMATT = 70.39 \\
            Route 1: Chosen 11 times, EWMATT = 72.00 \\
            Route 2: Chosen 2 times, EWMATT = 73.68 \\
            Route 3: Chosen 9 times, EWMATT = 73.61 \\
            Route 4: Chosen 7 times, EWMATT = 72.04
        \end{tabular} & 
        Route 2 & 
        \begin{tabular}[c]{@{}c@{}}
            Exploring route 2 despite its higher EWMATT \\
            allows for gathering more data and potentially \\
             discovering a new optimal route. \\
            The limited data on route 2 warrants further \\ 
            exploration to accurately assess its viability.
        \end{tabular} \\
        \midrule
        Exploit & 
        \begin{tabular}[c]{@{}c@{}}
            Route 0: Chosen 30 times, EWMATT = 59.06 \\
            Route 1: Chosen 44 times, EWMATT = 57.55 \\
            Route 2: Chosen 8 times, EWMATT = 61.21 \\
            Route 3: Chosen 5 times, EWMATT = 59.78 \\
            Route 4: Chosen 6 times, EWMATT = 60.95
        \end{tabular} & 
        Route 1 & 
        \begin{tabular}[c]{@{}c@{}}
            Route 1 has the lowest EWMATT (57.55) and \\
            has been  chosen the most frequently (44 times), \\
            suggesting it is a reliable and efficient route.
        \end{tabular} \\
        \bottomrule
    \end{tabular}
\end{table}

\subsubsection{Individual-level behavior of LLMTravelers}
Figure~\ref{fig:individual_choices} shows the route choice evolution for two LLM-gpt35-based agents in the OW network, traveling from Node 1 to Node 12. In the initial days, "LLM-gpt35 \#1" gradually explored from "Route 0" to "Route 4", while "LLM-gpt35 \#2" explored in the opposite direction, transitioning from "Route 4" to "Route 0". The daily travel time for each route fluctuated, leading to differences in the agents' learning and exploration outcomes. Over time, "LLM-gpt35 \#1" learned that "Route 2" was optimal and consistently chose it after 50 days. In contrast, "LLM-gpt35 \#2" identified "Route 1" as the best choice. These varied experiences, stored in memory, shaped their EWMATT, leading to different route choice behaviors.

Figure~\ref{fig:Comparison_of_two_agents} presents the two agents' retrieved memories after 100 days, showing their chosen times and EWMATT for different routes. In Figure~\ref{fig:Comparison_of_two_agents}(a), the EWMATT differences between the two agents for each route were within 1.5 minutes. However, their preferences diverged. For "LLM-gpt35 \#1," "Route 1" had the lowest EWMATT, followed by "Route 2". For "LLM-gpt35 \#2," "Route 2" was the best, followed by "Route 0." Figure~\ref{fig:Comparison_of_two_agents}(b) reveals significant differences in the chosen times for each route, especially "Routes 1" and "Route 2". Despite this, all routes were explored at least six times, reflecting exploratory behavior often seen in real-world travelers. These results highlight how memory and experience shape route choices, consistent with realistic decision-making patterns where individuals balance exploration and exploitation based on past experiences. Additionally, this behavior aligns with the prompt's guidance to "consider both well-traveled routes and those less explored."

\begin{figure}[h]
    \centering
    \includegraphics[width=0.98\textwidth]{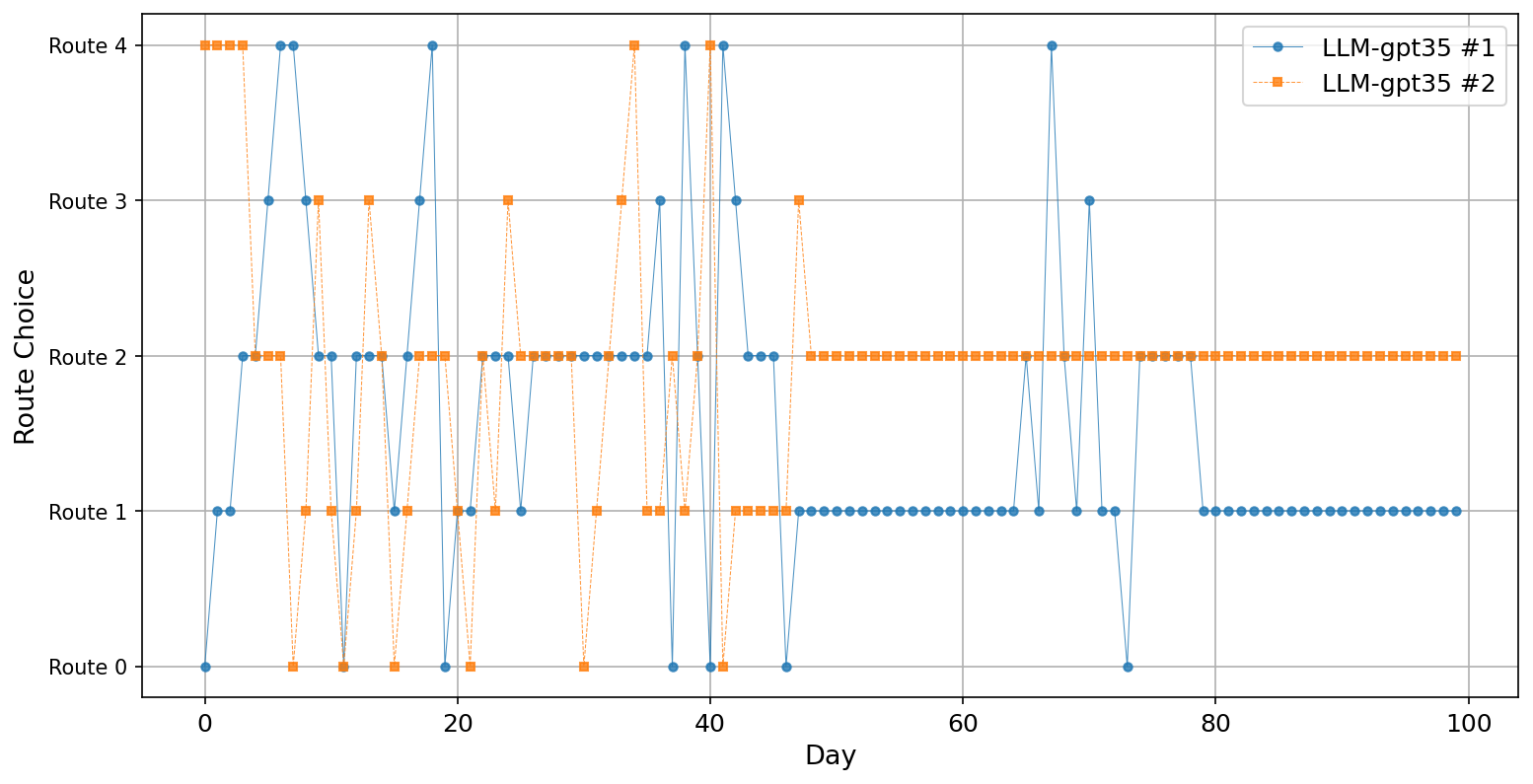}
    \caption{Evolution of route choices over days}
    \label{fig:individual_choices}
\end{figure}

\begin{figure}[h]
    \centering
    \includegraphics[width=0.98\textwidth]{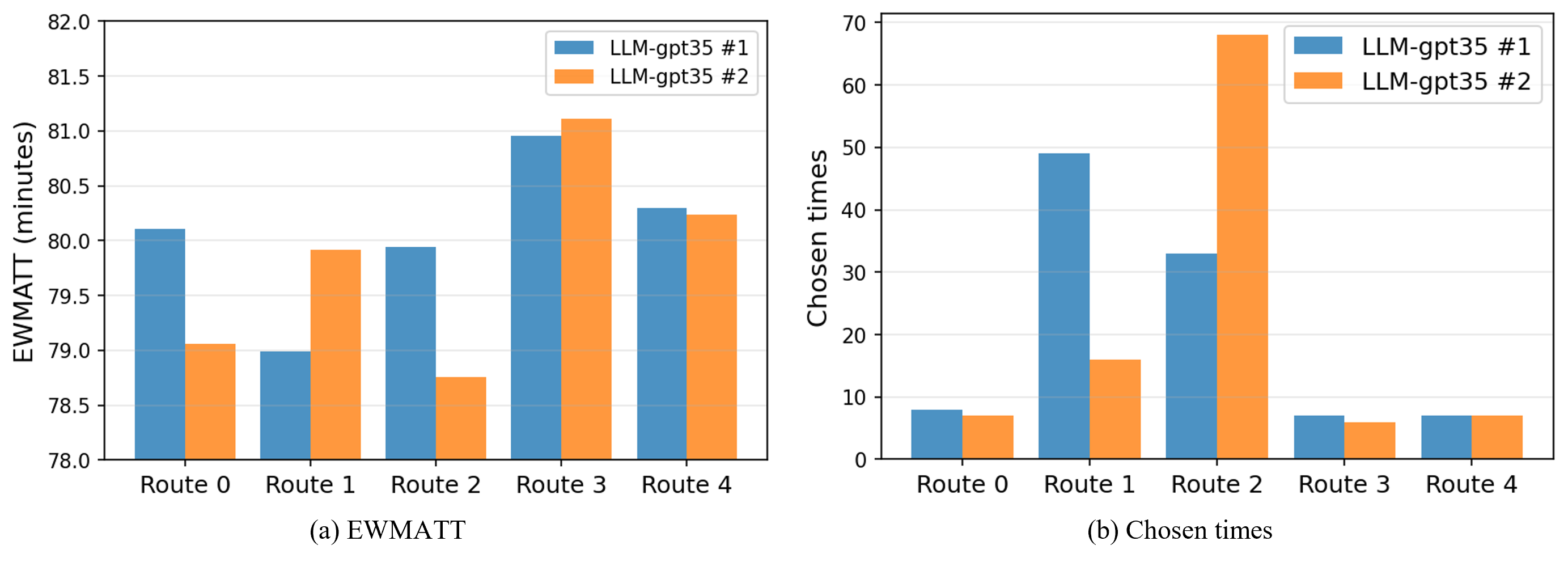}
    \caption{Learned memory of two LLMTravelers across routes}
    \label{fig:Comparison_of_two_agents}
\end{figure}

\subsubsection{Comparison with MNL and RL-based method}

Figure~\ref{fig:compare_with_MNL} shows that LLM-gpt35, MNL-0.3, and the RL-based agent converge to a similar average travel time of approximately 71.1 minutes. However, both LLM-gpt35 and the RL-based agent exhibit fluctuations around the UE even after convergence, achieving a slightly lower travel time, a behavior not observed in the MNL model. Furthermore, the MNL model demonstrates significantly faster convergence during the initial several days. This is because the MNL model shares the experience of all routes to calculate EWMATT and proportionally allocates choices, whereas the LLM-gpt35 optimizes choices individually based on accumulated experience. Notably, the RL-based agent requires substantially more data to achieve similar results, a reflection of the relatively low sample efficiency of reinforcement learning. For instance, the RL-based method needs nearly 4000 days of data to achieve results comparable to LLM-gpt35's convergence within 100 days. This highlights one of the key limitations of reinforcement learning in this context.

\begin{figure}[h]
    \centering
    \includegraphics[width=0.8\textwidth]{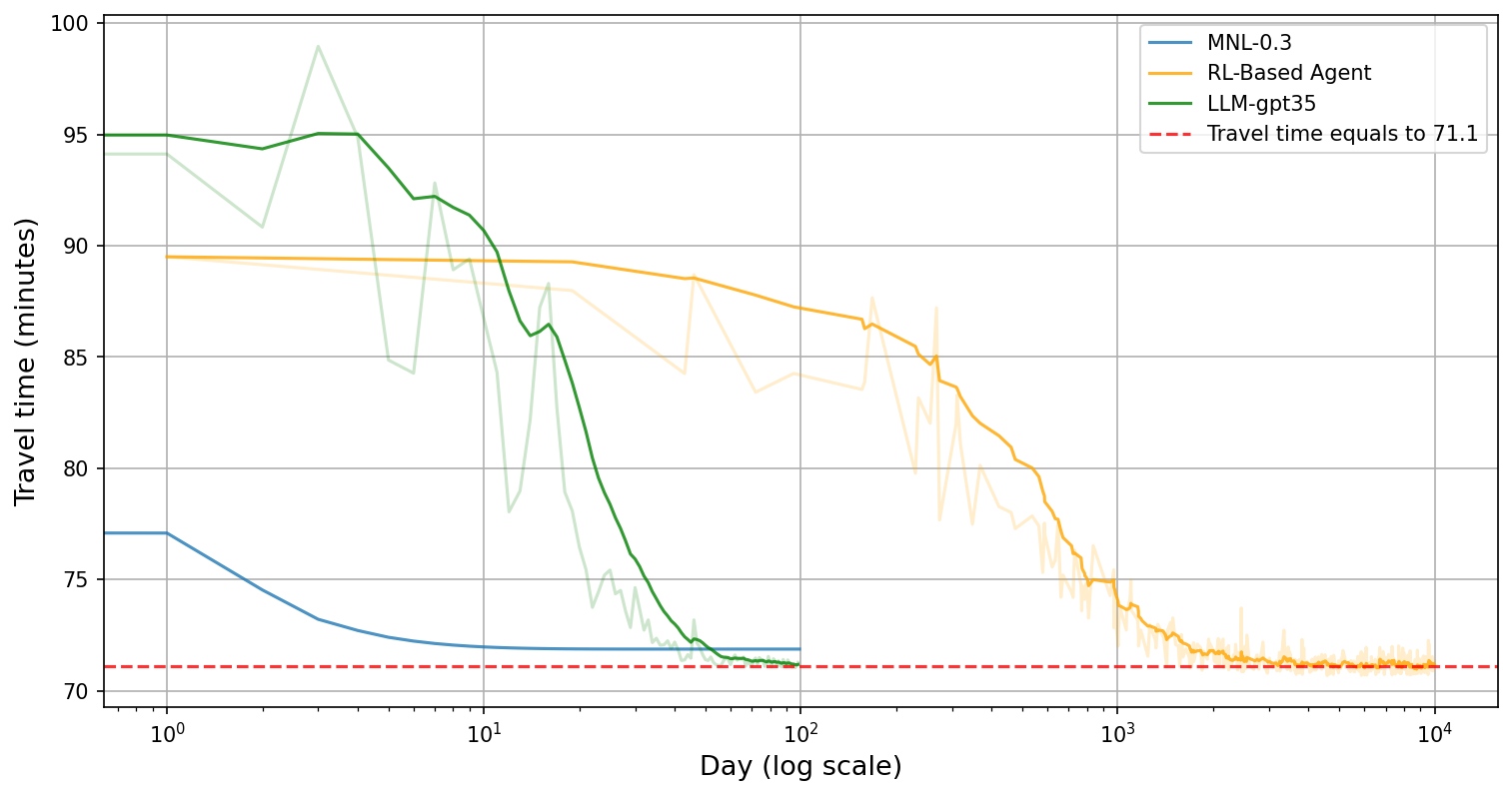}
    \caption{Travel time over days of different methods}
    \label{fig:compare_with_MNL}
\end{figure}

\subsubsection{Interpretability}

Table~\ref{tab:ow_route_choice_reason} and Table~\ref{tab:binary_choice_and_reason_examples} show that LLMTravelers typically provide reasonable explanations for their decisions. This reasoning process not only explains the choice but also reflects the model's thinking process. For the three examples in Table~\ref{tab:ow_route_choice_reason}, the LLMTravelers consider the current data, focusing particularly on the chosen times and EWMATT, and balance exploration and exploitation before making the final choice. These explanations align with the prompt's guidance to "think step by step," which encourages the agent to generate intermediate reasoning steps \cite{wei2022chain, zhang2022automatic}. Moreover, the variety of strategies shown in Table~\ref{tab:binary_choice_and_reason_examples} highlights their ability to mimic human decision-making. These agents combine rational analysis with uncertainty, making choices that reflect both historical experiences and personality traits. For instance, while some agents opt for route choices based on lower average travel times, others incorporate randomness or strategic considerations, such as balancing the impact of others’ choices. This combination of logic and unpredictability enhances the LLMTravelers' applicability for simulating route choice behaviors in transportation studies, providing both clarity and variability in their decision-making.

\section{Conclusion}
\label{sec:conclusion}

This study demonstrates the potential of LLMs as a novel approach to route choice modeling. The proposed framework, LLMTraveler, integrates an LLM with a memory system that allows the agent to interact with its environment, learn from past experiences, and adapt its decisions. The performance of LLMTraveler was evaluated in both single and multi-OD pair scenarios. In the single OD pair scenarios, the agent’s route-switching behavior aligns with most patterns observed in laboratory data, with some of these patterns not being fully explained by traditional models. In the multi-OD pair scenario, LLMTraveler replicates human-like behavior at both aggregate and individual levels. The route choice outcomes are comparable to those produced by traditional MNL models and RL-based agents. Additionally, LLMTraveler provides natural language explanations for its decisions, offering transparency and insight into its reasoning. The study also shows that lightweight, open-source LLMs can effectively replicate human-like route choice behavior, with only minor performance differences compared to larger, closed-source models.

However, this study is limited to route choice. Future research could extend this approach to other transportation decisions, such as mode choice and departure time. Additionally, calibrating LLMs with real human activity data could improve their alignment with actual behavior, enhancing their realism and applicability.

\section*{Acknowledgments}

The work was supported by start-up funds with No. MSRI8001004 and No. MSRI9002005, partly by the TRENoP research center fund at KTH, Sweden.

\section*{AUTHOR CONTRIBUTIONS}
The authors confirm contribution to the paper as follows: study conception and design: Z Ma, L Wang, Z He, C Lyu, P Duan, X Chen, N Zheng; methodology: P Duan, L Wang, Z Ma, Z He, C Lyu, L Yao; data collection: L Wang, C Lyu, X Chen; analysis and interpretation of results: L Wang, C Lyu, Z Ma, X Chen, P Duan, Z He; draft manuscript preparation: L Wang, P Duan, Z Ma, X Chen, C Lyu. manuscript revision: P Duan, Z Ma, Z He, X Chen, C Lyu, N Zheng, L Yao. All authors reviewed the results and approved the final version of the manuscript.

\bibliographystyle{unsrt}  
\bibliography{references}

\begin{thebibliography}{10}

\bibitem{prato2009route}
Carlo~Giacomo Prato.
\newblock Route choice modeling: past, present and future research directions.
\newblock {\em Journal of choice modelling}, 2(1):65--100, 2009.

\bibitem{he2012modeling}
Xiaozheng He and Henry~X Liu.
\newblock Modeling the day-to-day traffic evolution process after an unexpected network disruption.
\newblock {\em Transportation Research Part B: Methodological}, 46(1):50--71, 2012.

\bibitem{xiao2016physics}
Feng Xiao, Hai Yang, and Hongbo Ye.
\newblock Physics of day-to-day network flow dynamics.
\newblock {\em Transportation Research Part B: Methodological}, 86:86--103, 2016.

\bibitem{guo2011bounded}
Xiaolei Guo and Henry~X Liu.
\newblock Bounded rationality and irreversible network change.
\newblock {\em Transportation Research Part B: Methodological}, 45(10):1606--1618, 2011.

\bibitem{ye2017rational}
Hongbo Ye and Hai Yang.
\newblock Rational behavior adjustment process with boundedly rational user equilibrium.
\newblock {\em Transportation Science}, 51(3):968--980, 2017.

\bibitem{xu2011decision}
Hongli Xu, Jing Zhou, and Wei Xu.
\newblock A decision-making rule for modeling travelers’ route choice behavior based on cumulative prospect theory.
\newblock {\em Transportation Research Part C: Emerging Technologies}, 19(2):218--228, 2011.

\bibitem{wang2013combined}
Guangchao Wang, Shoufeng Ma, and Ning Jia.
\newblock A combined framework for modeling the evolution of traveler route choice under risk.
\newblock {\em Transportation Research Part C: Emerging Technologies}, 35:156--179, 2013.

\bibitem{kumar2015day}
Amit Kumar and Srinivas Peeta.
\newblock A day-to-day dynamical model for the evolution of path flows under disequilibrium of traffic networks with fixed demand.
\newblock {\em Transportation Research Part B: Methodological}, 80:235--256, 2015.

\bibitem{he2016marginal}
Xiaozheng He and Srinivas Peeta.
\newblock A marginal utility day-to-day traffic evolution model based on one-step strategic thinking.
\newblock {\em Transportation Research Part B: Methodological}, 84:237--255, 2016.

\bibitem{wei2016day}
Fangfang Wei, Ning Jia, and Shoufeng Ma.
\newblock Day-to-day traffic dynamics considering social interaction: from individual route choice behavior to a network flow model.
\newblock {\em Transportation Research Part B: Methodological}, 94:335--354, 2016.

\bibitem{smith1984stability}
Michael~J Smith.
\newblock The stability of a dynamic model of traffic assignment—an application of a method of lyapunov.
\newblock {\em Transportation science}, 18(3):245--252, 1984.

\bibitem{horowitz1984stability}
Joel~L Horowitz.
\newblock The stability of stochastic equilibrium in a two-link transportation network.
\newblock {\em Transportation Research Part B: Methodological}, 18(1):13--28, 1984.

\bibitem{iida1992experimental}
Yasunori Iida, Takamasa Akiyama, and Takashi Uchida.
\newblock Experimental analysis of dynamic route choice behavior.
\newblock {\em Transportation Research Part B: Methodological}, 26(1):17--32, 1992.

\bibitem{selten2007commuters}
Reinhard Selten, Thorsten Chmura, Thomas Pitz, Sebastian Kube, and Michael Schreckenberg.
\newblock Commuters route choice behaviour.
\newblock {\em Games and Economic Behavior}, 58(2):394--406, 2007.

\bibitem{meneguzzer2013day}
Claudio Meneguzzer and Alberto Olivieri.
\newblock Day-to-day traffic dynamics: laboratory-like experiment on route choice and route switching in a simple network with limited feedback information.
\newblock {\em Procedia-Social and Behavioral Sciences}, 87:44--59, 2013.

\bibitem{zou2013dynamic}
Mingqiao Zou, Xiqun~Michael Chen, Haixiao Yu, Yinan Tong, Ziwei Huang, Meng Li, and Haoda Zou.
\newblock Dynamic transportation planning and operations: concept, framework and applications in china.
\newblock {\em Procedia-Social and Behavioral Sciences}, 96:2332--2343, 2013.

\bibitem{jha1998perception}
Mithilesh Jha, Samer Madanat, and Srinivas Peeta.
\newblock Perception updating and day-to-day travel choice dynamics in traffic networks with information provision.
\newblock {\em Transportation Research Part C: Emerging Technologies}, 6(3):189--212, 1998.

\bibitem{nakayama2000route}
Shoichiro Nakayama and Ryuichi Kitamura.
\newblock Route choice model with inductive learning.
\newblock {\em Transportation Research Record}, 1725(1):63--70, 2000.

\bibitem{dia2002agent}
Hussein Dia.
\newblock An agent-based approach to modelling driver route choice behaviour under the influence of real-time information.
\newblock {\em Transportation Research Part C: Emerging Technologies}, 10(5-6):331--349, 2002.

\bibitem{rossetti2005dynamic}
Rosaldo~JF Rossetti and Ronghui Liu.
\newblock A dynamic network simulation model based on multi-agent systems.
\newblock In {\em Applications of Agent Technology in Traffic and Transportation}, pages 181--192. Springer, 2005.

\bibitem{ramos2018analysing}
Gabriel de~O Ramos, Ana~LC Bazzan, and Bruno~C da~Silva.
\newblock Analysing the impact of travel information for minimising the regret of route choice.
\newblock {\em Transportation Research Part C: Emerging Technologies}, 88:257--271, 2018.

\bibitem{mao2018reinforcement}
Chao Mao and Zuojun Shen.
\newblock A reinforcement learning framework for the adaptive routing problem in stochastic time-dependent network.
\newblock {\em Transportation Research Part C: Emerging Technologies}, 93:179--197, 2018.

\bibitem{zhou2020reinforcement}
Bo~Zhou, Qiankun Song, Zhenjiang Zhao, and Tangzhi Liu.
\newblock A reinforcement learning scheme for the equilibrium of the in-vehicle route choice problem based on congestion game.
\newblock {\em Applied Mathematics and Computation}, 371:124895, 2020.

\bibitem{shou2022multi}
Zhenyu Shou, Xu~Chen, Yongjie Fu, and Xuan Di.
\newblock Multi-agent reinforcement learning for markov routing games: A new modeling paradigm for dynamic traffic assignment.
\newblock {\em Transportation Research Part C: Emerging Technologies}, 137:103560, 2022.

\bibitem{achiam2023gpt}
Josh Achiam, Steven Adler, Sandhini Agarwal, Lama Ahmad, Ilge Akkaya, Florencia~Leoni Aleman, Diogo Almeida, Janko Altenschmidt, Sam Altman, Shyamal Anadkat, et~al.
\newblock Gpt-4 technical report.
\newblock {\em arXiv preprint arXiv:2303.08774}, 2023.

\bibitem{mei2024turing}
Qiaozhu Mei, Yutong Xie, Walter Yuan, and Matthew~O Jackson.
\newblock A turing test of whether ai chatbots are behaviorally similar to humans.
\newblock {\em Proceedings of the National Academy of Sciences}, 121(9):e2313925121, 2024.

\bibitem{guo2024large}
Taicheng Guo, Xiuying Chen, Yaqi Wang, Ruidi Chang, Shichao Pei, Nitesh~V Chawla, Olaf Wiest, and Xiangliang Zhang.
\newblock Large language model based multi-agents: A survey of progress and challenges.
\newblock {\em arXiv preprint arXiv:2402.01680}, 2024.

\bibitem{park2023generative}
Joon~Sung Park, Joseph O'Brien, Carrie~Jun Cai, Meredith~Ringel Morris, Percy Liang, and Michael~S Bernstein.
\newblock Generative agents: Interactive simulacra of human behavior.
\newblock In {\em Proceedings of the 36th annual acm symposium on user interface software and technology}, pages 1--22, 2023.

\bibitem{gao2024large}
Chen Gao, Xiaochong Lan, Nian Li, Yuan Yuan, Jingtao Ding, Zhilun Zhou, Fengli Xu, and Yong Li.
\newblock Large language models empowered agent-based modeling and simulation: A survey and perspectives.
\newblock {\em Humanities and Social Sciences Communications}, 11(1):1--24, 2024.

\bibitem{brown2020language}
Tom~B Brown.
\newblock Language models are few-shot learners.
\newblock {\em arXiv preprint arXiv:2005.14165}, 2020.

\bibitem{wang2024survey}
Lei Wang, Chen Ma, Xueyang Feng, Zeyu Zhang, Hao Yang, Jingsen Zhang, Zhiyuan Chen, Jiakai Tang, Xu~Chen, Yankai Lin, et~al.
\newblock A survey on large language model based autonomous agents.
\newblock {\em Frontiers of Computer Science}, 18(6):186345, 2024.

\bibitem{vaswani2017attention}
A~Vaswani.
\newblock Attention is all you need.
\newblock {\em Advances in Neural Information Processing Systems}, 2017.

\bibitem{madaan2024self}
Aman Madaan, Niket Tandon, Prakhar Gupta, Skyler Hallinan, Luyu Gao, Sarah Wiegreffe, Uri Alon, Nouha Dziri, Shrimai Prabhumoye, Yiming Yang, et~al.
\newblock Self-refine: Iterative refinement with self-feedback.
\newblock {\em Advances in Neural Information Processing Systems}, 36, 2024.

\bibitem{mwale2022factors}
Moses Mwale, Rose Luke, and Noleen Pisa.
\newblock Factors that affect travel behaviour in developing cities: A methodological review.
\newblock {\em Transportation Research Interdisciplinary Perspectives}, 16:100683, 2022.

\bibitem{parr2016differential}
Morgan~N Parr, Lesley~A Ross, Benjamin McManus, Haley~J Bishop, Shannon~MO Wittig, and Despina Stavrinos.
\newblock Differential impact of personality traits on distracted driving behaviors in teens and older adults.
\newblock {\em Accident Analysis \& Prevention}, 92:107--112, 2016.

\bibitem{vacca2019should}
Alessandro Vacca, Carlo~Giacomo Prato, and Italo Meloni.
\newblock Should i stay or should i go? investigating route switching behavior from revealed preferences data.
\newblock {\em Transportation}, 46:75--93, 2019.

\bibitem{wei2022chain}
Jason Wei, Xuezhi Wang, Dale Schuurmans, Maarten Bosma, Fei Xia, Ed~Chi, Quoc~V Le, Denny Zhou, et~al.
\newblock Chain-of-thought prompting elicits reasoning in large language models.
\newblock {\em Advances in neural information processing systems}, 35:24824--24837, 2022.

\bibitem{zhang2022automatic}
Zhuosheng Zhang, Aston Zhang, Mu~Li, and Alex Smola.
\newblock Automatic chain of thought prompting in large language models.
\newblock {\em arXiv preprint arXiv:2210.03493}, 2022.

\bibitem{yao2022react}
Shunyu Yao, Jeffrey Zhao, Dian Yu, Nan Du, Izhak Shafran, Karthik Narasimhan, and Yuan Cao.
\newblock React: Synergizing reasoning and acting in language models.
\newblock {\em arXiv preprint arXiv:2210.03629}, 2022.

\bibitem{yen1970algorithm}
Jin~Y Yen.
\newblock An algorithm for finding shortest routes from all source nodes to a given destination in general networks.
\newblock {\em Quarterly of applied mathematics}, 27(4):526--530, 1970.

\bibitem{qi2023investigating}
Hang Qi, Ning Jia, Xiaobo Qu, and Zhengbing He.
\newblock Investigating day-to-day route choices based on multi-scenario laboratory experiments, part i: Route-dependent attraction and its modeling.
\newblock {\em Transportation Research Part A: Policy and Practice}, 167:103553, 2023.

\bibitem{zhang2018cumulative}
Chong Zhang, Tian-Liang Liu, Hai-Jun Huang, and Jian Chen.
\newblock A cumulative prospect theory approach to commuters’ day-to-day route-choice modeling with friends’ travel information.
\newblock {\em Transportation Research Part C: Emerging Technologies}, 86:527--548, 2018.

\bibitem{meneguzzer2019contrarians}
Claudio Meneguzzer.
\newblock Contrarians do better: Testing participants’ response to information in a simulated day-to-day route choice experiment.
\newblock {\em Travel Behaviour and Society}, 15:146--156, 2019.

\bibitem{zhao2016experiment}
Chuan-Lin Zhao and Hai-Jun Huang.
\newblock Experiment of boundedly rational route choice behavior and the model under satisficing rule.
\newblock {\em Transportation Research Part C: Emerging Technologies}, 68:22--37, 2016.

\bibitem{qi2024investigating}
Hang Qi, Ning Jia, Xiaobo Qu, and Zhengbing He.
\newblock Investigating day-to-day route choices based on multi-scenario laboratory experiments, part ii: Route-dependent attraction-based stochastic process model.
\newblock {\em Communications in Transportation Research}, 4:100123, 2024.

\bibitem{zhang2001equivalence}
Ding Zhang, Anna Nagurney, and Jiahao Wu.
\newblock On the equivalence between stationary link flow patterns and traffic network equilibria.
\newblock {\em Transportation Research Part B: Methodological}, 35(8):731--748, 2001.

\bibitem{daganzo1977stochastic}
Carlos~F Daganzo and Yosef Sheffi.
\newblock On stochastic models of traffic assignment.
\newblock {\em Transportation science}, 11(3):253--274, 1977.

\bibitem{dubey2024llama}
Abhimanyu Dubey, Abhinav Jauhri, Abhinav Pandey, Abhishek Kadian, Ahmad Al-Dahle, Aiesha Letman, Akhil Mathur, Alan Schelten, Amy Yang, Angela Fan, et~al.
\newblock The llama 3 herd of models.
\newblock {\em arXiv preprint arXiv:2407.21783}, 2024.

\bibitem{young2024yi}
Alex Young, Bei Chen, Chao Li, Chengen Huang, Ge~Zhang, Guanwei Zhang, Heng Li, Jiangcheng Zhu, Jianqun Chen, Jing Chang, et~al.
\newblock Yi: Open foundation models by 01. ai.
\newblock {\em arXiv preprint arXiv:2403.04652}, 2024.

\bibitem{dixit2014equilibrium}
Vinayak~V Dixit and Laurent Denant-Boemont.
\newblock Is equilibrium in transport pure nash, mixed or stochastic?
\newblock {\em Transportation Research Part C: Emerging Technologies}, 48:301--310, 2014.

\bibitem{de2024modelling}
Juan de~Dios~Ort{\'u}zar and Luis~G Willumsen.
\newblock {\em Modelling transport}.
\newblock John wiley \& sons, 2024.

\bibitem{yu2022surprising}
Chao Yu, Akash Velu, Eugene Vinitsky, Jiaxuan Gao, Yu~Wang, Alexandre Bayen, and Yi~Wu.
\newblock The surprising effectiveness of ppo in cooperative multi-agent games.
\newblock {\em Advances in Neural Information Processing Systems}, 35:24611--24624, 2022.

\bibitem{kadavath2022language}
Saurav Kadavath, Tom Conerly, Amanda Askell, Tom Henighan, Dawn Drain, Ethan Perez, Nicholas Schiefer, Zac Hatfield-Dodds, Nova DasSarma, Eli Tran-Johnson, et~al.
\newblock Language models (mostly) know what they know.
\newblock {\em arXiv preprint arXiv:2207.05221}, 2022.

\bibitem{leng2024taming}
Jixuan Leng, Chengsong Huang, Banghua Zhu, and Jiaxin Huang.
\newblock Taming overconfidence in llms: Reward calibration in rlhf.
\newblock {\em arXiv preprint arXiv:2410.09724}, 2024.

\end{thebibliography}

\end{document}